\documentclass[10pt,twocolumn,letterpaper]{article}

\usepackage{cvpr}              
\newcommand{\second}[1]{\textcolor{blue}{#1}}
\usepackage{algorithm,algpseudocode,amsmath}

\usepackage{xcolor,colortbl,array}
\definecolor{reconcol}{gray}{0.93}   
\definecolor{uncertcol}{gray}{0.97}  
\newcolumntype{R}{>{\columncolor{reconcol}}r}
\newcolumntype{U}{>{\columncolor{uncertcol}}r}
\definecolor{cvprblue}{rgb}{0.21,0.49,0.74}
\usepackage[pagebackref,breaklinks,colorlinks,allcolors=cvprblue]{hyperref}

\def\paperID{11487} 
\def\confName{CVPR}
\def\confYear{2026}

\title{Uncertainty Quantification in HSI Reconstruction using Physics-Aware Diffusion Priors and Optics-Encoded Measurements}

\author{
    Juan Romero$^{1}$ \quad
    Qiang Fu$^{1}$ \quad
    Matteo Ravasi$^{2}$ \quad
    Wolfgang Heidrich$^{1}$\\[0.5em]
    $^{1}$King Abdullah University of Science and Technology (KAUST), Thuwal, Saudi Arabia\\
    $^{2}$Shearwater Geoservices, Gatwick, United Kingdom\\[0.5em]
    {\tt\small juan.romeromurcia@kaust.edu.sa}
}

\begin{document}
\maketitle

\begin{abstract}
Hyperspectral image reconstruction from a compressed measurement is a highly ill-posed inverse problem. Current data-driven methods suffer from hallucination due to the lack of spectral diversity in existing hyperspectral image datasets, particularly when they are evaluated for the metamerism phenomenon. In this work, we formulate hyperspectral image (HSI) reconstruction as a Bayesian inference problem and propose a framework, HSDiff, that utilizes an unconditionally trained, pixel-level diffusion prior and posterior diffusion sampling to generate diverse HSI samples consistent with the measurements of various hyperspectral image formation models. We propose an enhanced metameric augmentation technique using region-based metameric black and partition-of-union spectral upsampling to expand training with physically valid metameric spectra, strengthening the prior diversity and improving uncertainty calibration. We utilize HSDiff to investigate how the studied forward models shape the posterior distribution and demonstrate that guiding with effective spectral encoding provides calibrated informative uncertainty compared to non-encoded models. Through the lens of the Bayesian framework, HSDiff offers a complete, high-performance method for uncertainty-aware HSI reconstruction. Our results also reiterate the significance of effective spectral encoding in snapshot hyperspectral imaging.
\end{abstract}   
\vspace{-0.5cm}
\section{Introduction}
\label{sec:intro}

Hyperspectral images (HSI) capture scene information across spectral bands, providing an augmented data cube beyond the three channels of human vision.  This dense spectral information is an essential component in various scientific and industrial fields, enabling applications from precision agriculture~\cite{lu_recent_2020} and non-invasive medical diagnostics~\cite{khan_modern_2018} to remote sensing~\cite{bioucas-dias_hyperspectral_2013} and the technical analysis of art~\cite{cucci_reflectance_2016}. However, the acquisition of this data remains a significant bottleneck. Specialized hyperspectral sensors are typically expensive, bulky, and suffer from trade-offs in spatial, spectral, and temporal resolution compared to standard RGB cameras~\cite{lodhi_hyperspectral_2019}. This gap between the value of HSI data and the difficulty of its acquisition motivates a strong demand for computational methods that can reconstruct or ``see'' the spectral world using ubiquitous, low-cost RGB sensors.

Recent advances in deep learning, particularly with Convolutional Neural Networks (CNNs) and Transformers, have significantly improved the quality of HSI reconstructions, achieving impressive scores on standard metrics such as Peak Signal-to-Noise Ratio (PSNR) and Spectral Angle Mapper (SAM)~\cite{huang_spectral_2022, cai_mst_2022, arad_ntire_2022}. However, this paradigm is built on a precarious foundation. As highlighted by recent work~\cite{fu_limitations_2025}, the public HSI datasets used for training severely lack diversity, especially in metameric colors. This scarcity means that state-of-the-art methods suffer from atypical overfitting, where models learn to fit small datasets but fail to generalize and lack robustness in the presence of real-world metamers~\cite{palmer_vision_1999,finlayson_metamer_2005}, noise, or even compression artifacts. Compounding this data-centric issue is a methodological one: most existing methods are exclusively deterministic. They learn a complex mapping that provides only a single point estimate for the latent HSI. This single-point-estimate approach is doubly problematic, as it not only ignores the ill-posed nature of the problem but also the estimate itself is trained on a metamer-poor dataset that hides the true scale of the solution's ambiguity. Therefore, the reported high scores in the literature appear overly optimistic and deserve a second, more critical examination.

\begin{figure*}
        \centering
        \includegraphics[width=\linewidth]{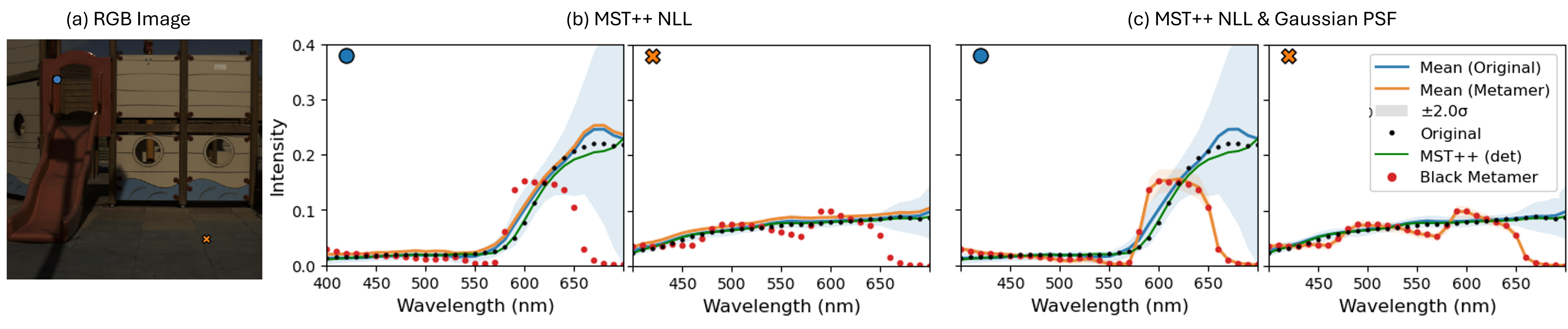}
        \caption{(a) RGB image with two marked pixel locations. (b) Predicted spectra at the marked pixels from MST++ trained (NLL-trained) without aberrations. The metamer prediction (orange solid) is completely wrong from the truth (red dotted). Moreover, the true metamer spectrum falls outside the confidence interval for the prediction, indicating that the prediction is ``confidently wrong''. (c) Predicted spectra at the same pixels from MST++ (NLL-trained) with optics-induced aberrations. Both original and metamer spectra are predicted correctly with reasonable uncertainty (mean $\pm \ 2 \ \sigma $). In both plots, the green line shows the standard deterministic MST++ prediction as a reference. 
        }
        \label{fig:motivation}
    \end{figure*}

This metamer-breaking effect is evident when comparing a state-of-the-art deterministic baseline MST++~\cite{cai_mst_2022}. As shown in Fig.~\ref{fig:motivation}, we train the model probabilistically using a negative log-likelihood (NLL) loss, with (Fig.~\ref{fig:motivation}c) and without (Fig.~\ref{fig:motivation}b)) optically aberrated measurements. We plot two spectra from the same spatial location for a pair of metamers that yield the same RGB. With NLL training on non-encoded measurements (Fig.~\ref{fig:motivation}b), MST++’s uncertainty bands largely cover variations around the one spectrum but fail to represent its metamer counterpart, because the measurement provides no lever to distinguish them when the inputs are the same. In other words, the trained model is confidently wrong for the metamer spectrum. In contrast, when the model is trained with optics-aware encoding (Fig.~\ref{fig:motivation}c), the wavelength-dependent PSF alters the target RGBs, so the two spectra map to measurably different observations. The network now separates them cleanly: predictions align with both the original and the metamer (as appropriate), and the uncertainty reflects reduced ambiguity where the PSF provides stronger spectral fingerprints. More details on this experiment can be found in the Supplementary Material.

In this work, we reframe HSI reconstruction as a Bayesian inference problem, where the goal is to characterize the full posterior distribution. For the prior distribution, we develop an unconditional diffusion model trained on an aggregation of the ARAD1K~\cite{arad_ntire_2022} and KAUST~\cite{li2021multispectral} datasets, and validate it on the ICVL \cite{arad_sparse_2016} and CAVE \cite{yasuma_generalized_2010} datasets. To address the known lack of metameric diversity in these datasets, we employ a dual augmentation strategy. We first adapt the Partition of Unity basis \cite{belcour_one--many_2023} to the sensor's RGB space, and secondly, we introduce a novel, texture-guided method for spatially-varying black metamer generation. This augmented prior and the physical forward model are integrated within a diffusion-based posterior sampling framework, which is then used to conduct a systematic analysis of how the forward operator impacts the geometry of the posterior distribution. By drawing multiple, high-fidelity posterior samples, the sample mean serves as the final reconstruction, and the sample variance provides a quantitative, pixel-wise measure of uncertainty. We investigate three forward operators, including a spectral-to-RGB projection without optical encoding, a spatio-spectral transform that modulates spectral point spread functions (PSFs), and a coded-aperture snapshot spectral imaging (CASSI) technique. Building on recent work~\cite{fu_limitations_2025} suggesting optical aberrations and encoded measurements are beneficial, we test this hypothesis from an uncertainty perspective. We demonstrate that guiding the sampler with optics-aware operators provides a more robust uncertainty quantification compared to non-encoded models. This establishes a direct connection between the fidelity of the physical model and the inherent ambiguity of the problem. Our primary contributions are:

\begin{itemize}
    \item We introduce a texture-guided metameric augmentation method based on metameric black and partition-of-union upsampling that enriches the training distribution, producing a robust diffusion prior that is aware of the forward operator's null-space.

    \item We propose ``HSDiff'', a framework based on efficient diffusion sampling that provides both high-quality reconstruction and a robust, quantitative, pixel-wise uncertainty map.

    \item We provide a systematic demonstration that guiding the diffusion process using physics-aware forward models with effective spectral encoding provided calibrated posterior variance, thereby quantifiably constraining the ill-posed solution space.

\end{itemize}

\section{Related work}
\label{sec:related}
\subsection{RGB-to-HSI Reconstruction}
    Hyperspectral reconstruction from RGB images has evolved rapidly over the last two decades, originating from traditional, prior-based approaches rooted in image statistics and constrained optimization frameworks~\cite{zhang_survey_2022}. Early efforts focused on using hand-crafted priors, such as spectral sparsity, spatial structure similarity, and inter-band correlation, to regularize the inherently ill-posed RGB-to-HSI inversion process~\cite{akhtar_hyperspectral_2020, fu_spectral_2018, li_locally_2018}. These methods aim to recover plausible hyperspectral information by exploiting known relationships among hyperspectral channels and leveraging the physics of image formation. While these classical strategies contribute towards making hyperspectral imaging more accessible in cost-sensitive scenarios, their performance is limited by reliance on manually crafted priors and lack of scalability to complex real-world scenes~\cite{zhang_survey_2022}.
    
    In recent years, the rise of deep learning has triggered data-driven hyperspectral reconstruction, with neural networks dominating state-of-the-art performance~\cite{huang_spectral_2022, cai_mst_2022, arad_ntire_2022, ahmed_comparative_2024, gao_deep-learning-based_2021}. These methods leverage paired RGB-HSI datasets to learn nonlinear mappings directly from data, utilizing architectures such as CNNs, residual networks, and transformer-inspired models. More sophisticated methods incorporate multi-scale feature extraction and spatial-spectral attention mechanisms. However, a recent study has revealed fundamental limitations of RGB-to-HSI~\cite{fu_limitations_2025}, showing that current datasets lack sufficient spectral diversity and severely under-represent metameric phenomena. In the absence of principled uncertainty quantification, models trained under such conditions can yield confidently wrong outputs, especially when encountering metameric ambiguities that they were never adequately trained on.


\subsection{Generative Models for Inverse Problems}

    Deterministic point estimates obscure ambiguity in ill-posed inverse problems, motivating generative approaches that approximate the full posterior over plausible solutions~\cite{kutyniok_deep_2022}. Rather than returning a single reconstruction, the objective is to characterize the set of images consistent with both the measurements and natural-image statistics. Early generative priors, such as GANs~\cite{goodfellow_generative_2020} and VAEs~\cite{kingma_auto-encoding_2022}, demonstrated promise but suffered from training instability and mode collapse (GANs) or over-smoothing and limited fidelity (VAEs). Denoising diffusion models (DDPMs)~\cite{ho_denoising_2020} address these issues by learning to reverse a multi-step noising process, yielding stable training and state-of-the-art sample quality. There are multiple ways to incorporate diffusion into inverse problems, and several taxonomies have been proposed~\cite{daras_survey_2024,chung_diffusion_2025}. In a task-agnostic framing centered on whether we retrain and how we enforce data consistency, existing methods fall into three broad groups:

    \textbf{(i) Task-specific conditioning / supervised.} One retrains or fine-tunes a conditional diffusion model to generate solutions directly conditioned on the measurement and, optionally, physics tokens or a pseudo-inverse. This pathway offers high throughput at test time but typically requires re-training per task/operator or camera model~\cite{li_srdiff_2021,saharia_palette_2022,whang_deblurring_2021}.
    
    \textbf{(ii) Unconditional prior with sampling-time guidance / unsupervised.} One keeps an unconditional prior and enforces data consistency during sampling via likelihood-guided updates. Examples include DPS, which injects a likelihood gradient into the score field~\cite{chung_diffusion_2024}, and PGDM, which uses pseudo-inverse or projection-style guidance~\cite{song_pseudoinverse-guided_2022}; related approaches adapt guidance weights, proximal corrections, or noise schedules to balance fidelity and realism~\cite{meng_diffusion_2024,wu_diffusion_2024,kawar_denoising_2022}.
    
    \textbf{(iii) Unrolled / proximal hybrids.} The score (or denoiser) serves as a proximal operator within an iterative solver (e.g., half-quadratic splitting, PnP-ADMM), trading some modularity for additional algorithmic control and potential convergence guarantees under assumptions on the denoiser~\cite{wu_principled_2024,zheng_inversebench_2025,xu_provably_nodate}.
    
    In this work, we adopt a DPS-style strategy augmented with a perturbed likelihood from score-based data assimilation that attenuates guidance at high noise levels~\cite{rozet_score-based_2023, chakraborty_multimodal_2025}. This family of methods avoids paired supervision and adapts to diverse operators, linear or nonlinear, noisy or noise-free, via automatic differentiation of the physics, enabling practical application across settings such as super-resolution~\cite{gendy_diffusion_2025}, phase retrieval~\cite{shoushtari_dolph_2022}, seismic inversion~\cite{ravasi_geophysical_2025} without retraining the prior. Crucially, by decoupling the learned prior from the measurement model, DPS-like approaches allow us to compare and evaluate different data-modeling operators within a single posterior-sampling framework. In the HSI context, diffusion models have been widely used for super resolution \cite{wu_hsr-diff_2023}, classification \cite{chen_spectraldiff_2023}, and reconstruction \cite{pang_hir-diff_2024,hazineh_grayscale_2025}. 

\subsection{Optical Encoding}

    Beyond pixel-wise spectral mixing through the spectral response function (SRF), optics can be engineered to encode wavelength information into space via a wavelength-dependent point-spread function (PSF)~\cite{jeon_compact_2019,cao_prism-mask_2011,baek_compact_2017,hazineh2024grayscale}. Dispersive elements (e.g., gratings, DOEs, metalens) induce controlled, band-specific blurs, shifts, rotations, or shape changes so that each wavelength leaves a distinct spatial "fingerprint". Intuitively, this converts some of the purely spectral ambiguity into measurable spatial structures: two metameric spectra under a linear projection model would produce different images after passing through a spectrally-encoded forward model. In linear terms, the forward operator gains band-dependent diversity, increasing its effective rank and reducing coherence between columns. The inverse problem then becomes better conditioned and less ill-posed.

    Another class of optical encoding introduces spatial-spectral encoding in the CASSI architecture~\cite{wagadarikar2008single,DeepCASSI:SIGA:2017,meng2020end}. An objective lens is first used to form an intermediate image plane, on which a binary random mask is placed to impose a spatial encoding. A relay system with a dispersive optical element (e.g., double Amici prism) is then used to spectrally spread out the spectra in one direction on the image sensor. The combination of the spatial mask and the spectral dispersion offers an encoding scheme for the subsequent neural network to reconstruct the hyperspectral image from the encoded measurement.

    In this paper, we evaluate all these existing optical encoding methods through our uncertainty-aware framework to reassess their effectiveness in encoding. The results offer insights into how to evaluate the effectiveness of the spectral encoding and inspire new encoding designs.

\section{Methodology}
\label{sec:methodology}

\subsection{Bayesian Spectral Image Formation Model}

    The RGB measurement is a linear spectral mixing of the hyperspectral image (HSI). Let $\mathbf{X}\in\mathbb{R}^{MN\times K}$ be the HSI reshaped to pixels-by-bands and $\mathbf{Y}\in\mathbb{R}^{MN\times 3}$ the RGB image. With camera spectral response function (SRF) $\mathbf{Q}\in\mathbb{R}^{K\times 3}$,
    \begin{equation}
        \mathbf{Y}=\mathbf{X}\mathbf{Q}.
    \end{equation}    
    This mapping is severely ill-posed: distinct spectra $\mathbf{X}_1,\mathbf{X}_2$ can yield the same RGB, a phenomenon known as metamerism. In practice, optics introduce wavelength-dependent blur (chromatic aberration), which we model with a per-band point-spread function (PSF). Let $\mathbf{H}_k\in\mathbb{R}^{MN\times MN}$ be the convolution matrix for band $k$, and define the operator $\mathcal{H}(\mathbf{X})=[\mathbf{H}_1\mathbf{x}_1,\ldots,\mathbf{H}_K\mathbf{x}_K,]\in\mathbb{R}^{MN\times K}$, where $\mathbf{x}_k$ is the $k$-th column of $\mathbf{X}$. The optics-aware forward model~\cite{fu_limitations_2025} is then
    \begin{equation}
        \mathbf{Y}=\mathcal{H}(\mathbf{X})\mathbf{Q}.
    \end{equation}
    Far from being a nuisance, this spatio-spectral entanglement provides wavelength-specific spatial fingerprints that help disambiguate metamers and better constrain the inverse problem. 

    The CASSI system operates in a slightly different way, where the measurement is sum of spectrally sheared and masked spectral images~\cite{meng2020end},
    \begin{equation}
        \mathbf{Y} = \sum_{i=1}^{K} \mathbf{M}^*_i \mathbf{x}^*_i = \mathcal{C} \left(\mathbf{X}\right),
    \end{equation}
    where $\mathbf{M}^*_i$ is the sheared mask, $\mathbf{x}^*_i$ is the sheared spectral image in the $i$-band, and $\mathcal{C}$ is the corresponding operator. 
    
    Building on this optics-aware model, we adopt a Bayesian view. Given $\mathbf{Y}=\mathcal{A}(\mathbf{X})+\boldsymbol{\epsilon}$ with $\mathcal{A}$ as the linear operator $\mathcal{A}(\mathbf{X}) := \mathcal{H}(\mathbf{X})$ and $\mathcal{A}(\mathbf{X}) := \mathcal{C}(\mathbf{X})$ for CASSI, and $\boldsymbol{\epsilon}\sim\mathcal{N}(0,\sigma_y^2\mathbf{I})$ as additive Gaussian noise, the likelihood is
    \begin{equation}
        p(\mathbf{Y}\mid \mathbf{X},\phi)\propto \exp\Big(-\tfrac{1}{2\sigma_y^2}\Vert\mathbf{Y}-\mathcal{A}(\mathbf{X})\Vert_2^2\Big),
    \end{equation}
    where $\phi=\{\mathcal{A},\sigma_y^2\}$ collects forward and noise parameters. We use a generative prior $p_\theta(\mathbf{X})$ over natural HSIs, instantiated by an unconditional diffusion model trained on spectra and spatial statistics. The posterior
    \begin{equation}
        p(\mathbf{X}\mid \mathbf{Y},\phi)\ \propto\ p(\mathbf{Y}\mid \mathbf{X},\phi)\ p_\theta(\mathbf{X}),
    \end{equation}
    balances data consistency with the learned HSI prior. An MAP estimation arises from maximizing this density, while our inference utilizes diffusion posterior sampling (DPS) to draw diverse $\mathbf{X}$ values consistent with $\mathbf{Y}$ and the physics. This yields calibrated uncertainty (e.g., per-band credible intervals and posterior variance maps), enabling principled comparisons across spectral encoders via changes in $\phi$. 

\subsection{A Diffusion Prior for Hyperspectral Images}
    
    We model the distribution of natural HSIs with a generative prior $p_\theta(\mathbf{X})$. Let $\mathbf{x}\in\mathbb{R}^{d}$ denote a vectorized HSI sample (e.g., $d{=}KMN)$. Following score-based diffusion \cite{song_score-based_2021} , the forward noising process evolves clean data $\mathbf{x}_0 \sim p(\mathbf{x})$ into noise via the stochastic differential equations (SDE)
    \begin{equation}
        \mathrm{d}\mathbf{x}_t = f(t)\mathbf{x}_t\mathrm{d}t + g(t)\mathrm{d}\mathbf{w}_t,\qquad t\in[0,T],
    \end{equation}
    where $f(t)$  is the drift (scales/contracts the signal as noise varies),  $g(t)$ controls the noise injection rate or diffusion strength, and $\mathbf{w}$ is the standard Wiener process. This forward process admits an exact reverse-time SDE~\cite{anderson_reverse-time_1982}:
    \begin{equation}
        \mathrm{d}\mathbf{x}_t = \big[f(t)\mathbf{x}_t - g(t)^2 \nabla{\mathbf{x}_t}\log p_t(\mathbf{x}_t)\big]\mathrm{d}t + g(t)\mathrm{d}\bar{\mathbf{w}}_t,
    \end{equation}
    which becomes generative when integrated from $t{=}T$ (pure noise) to $t{=}0$ using the score $\nabla{\mathbf{x}_t}\log p_t(\mathbf{x}_t)$. Intuitively, the score is a vector field over $\mathbf{x}_t$ that guides samples back to the data manifold as noise anneals. The learning task is therefore to approximate this score. In EDM \cite{karras_elucidating_2022}, a denoiser $D_\theta(\mathbf{x};\sigma)$ is parametrized so that it predicts the clean signal $\mathbf{x}_0$ from a noisy input $\mathbf{x}$ corrupted at a noise level $\sigma$. This yields the score proxy
    \begin{equation}
            \nabla_{\mathbf{x}}\log p_\sigma(\mathbf{x}) \approx \frac{D_\theta(\mathbf{x};\sigma)-\mathbf{x}}{\sigma^2},
    \end{equation}
    and a preconditioned regression objective
    \begin{equation}
        \mathcal{L}(\theta)=\mathbb{E}_{\sigma,\mathbf{x}_0,\boldsymbol{\epsilon}}\left[w(\sigma),\Vert D_\theta(\mathbf{x}_0+\sigma\boldsymbol{\epsilon};\sigma)-\mathbf{x}_0\Vert_2^2\right],
    \end{equation}
    where $\boldsymbol{\epsilon}\sim\mathcal{N}(\mathbf{0},\mathbf{I})$ and $w(\sigma)$ is a noise-level weighting that aligns the training geometry with the corruption process. Once trained, samples are generated by numerically integrating the reverse dynamics. EDM provides efficient few-step samplers (both deterministic and stochastic), and we adopt the stochastic variant as the backbone for posterior-guided inference.
        
    We parameterize the denoiser $D_\theta$ with a 2D U-Net that follows the ADM~\cite{dhariwal_diffusion_2021} design (residual blocks, skip connections, and time/noise conditioning with low-res attention), adapted to hyperspectral inputs of $256\times256\times31$. We prepend a spectral mixing stem: a 1D conv along the 31-band axis (kernel=5) to encourage local spectral smoothness, followed by a squeeze-and-excitation block that reweights bands using global context. The U-Net then processes spatial feature maps through a 256-128-64-32 pyramid (and symmetric up-path); we insert spatial self-attention at the $64\times64$ and $32\times32$ stages in both encoder and decoder to capture long-range dependencies. For the time/noise embedding, we use sinusoidal Fourier features of $\log \sigma$ (EDM’s $\sigma$-space) passed through a two-layer MLP; the resulting conditioning vector modulates each residual block via adaptive normalization, mirroring ADM’s conditioning pathway but with EDM preconditioning and loss weighting in place of the original ADM objective. The network predicts the clean target $\mathbf{x}_0$ at the input resolution, and we maintain an EMA of weights for sampling.

\subsection{Posterior Sampling and Uncertainty Quantification}
   
    The unconditional reverse-time SDE generates prior samples $p_{\theta}(\mathbf{x})$. To sample from the posterior $p(\mathbf{x}\mid \mathbf{y})$, we replace the prior score with the posterior score in the dynamics:
    \begin{equation}
        \mathrm{d}\mathbf{x}_t =\Big[f(t)\mathbf{x}t - g(t)^2,\nabla{\mathbf{x}_t}\log p_t(\mathbf{x}_t\mid \mathbf{y})\Big]\mathrm{d}t + g(t),\mathrm{d}\bar{\mathbf{w}}t .
    \end{equation}
    By Bayes’ rule, the posterior score decomposes as
    \begin{equation}
      \nabla{\mathbf{x}_t}\log p_t(\mathbf{x}_t\mid \mathbf{y})
      =\underbrace{\nabla{\mathbf{x}_t}\log p_t(\mathbf{x}_t)}_{\text{prior score}}
      +\underbrace{\nabla{\mathbf{x}_t}\log p(\mathbf{y}\mid \mathbf{x}_t)}_{\text{likelihood score}}.
    \end{equation}
    The prior score is provided by the denoiser $D_\theta$ and the likelihood score enforces data consistency with an observation or measurement $\mathbf{y}$. A variety of approaches have been proposed to compute the likelihood score and its integration with the prior score \cite{chung_diffusion_2024, song_pseudoinverse-guided_2022, wu_diffusion_2024}. In this work, we adopt the perturbed likelihood from score-based data assimilation (SDA) formalism \cite{rozet_score-based_2023}, and its extension to EDM \cite{chakraborty_multimodal_2025} that accounts for the variance in $\mathbf{x}_0$:

    \begin{equation}
        \label{eqn:likelihood}
        p(\mathbf{y}\mid \mathbf{x}_t)\approx \mathcal{N}\left(\mathbf{y}\ \middle|\ \mathcal{A}(\hat{\mathbf{x}}_{0}); \
        \Sigma_y +\frac{\sigma(t)^2}{\mu(t)^2}J \Gamma J^{\top}\right),
    \end{equation}
    where $\hat{\mathbf{x}}_{0}=D_{\theta}(\mathbf{x}_{t};\,\sigma(t))$ is the denoiser’s estimate of the clean HSI at noise level $\sigma(t)$, $\boldsymbol{\Sigma}_{y}$ is the measurement-noise covariance, $J=\left.\partial_{\mathbf{x}}\mathcal{A}\right|_{\hat{\mathbf{x}}_{0}}$ is the Jacobian of $\mathcal{A}$ evaluated at $\hat{\mathbf{x}}_{0}$, $\boldsymbol{\Gamma}$ approximates the covariance of $\mathbf{x}_{t}$, and $\mu(t)$ is the input scaling at diffusion time $t$. Following \cite{rozet_score-based_2023,chakraborty_multimodal_2025}, we approximate both $\boldsymbol{\Sigma}_{y}$ and $J\,\boldsymbol{\Gamma}\,J^{\top}$ with diagonal matrices. See supplementary material for details.

    We quantify \emph{pixel uncertainty} per spatial location and spectral band. Running the guided sampler $N$ times from i.i.d. noise $\{\mathbf{x}_T^{(k)}\}_{k=1}^{N}$ yields samples $\{\mathbf{x}_0^{(k)}\}_{k=1}^{N}$ with $\mathbf{x}_0^{(k)}\in\mathbb{R}^{M\times N\times K}$. The posterior mean (point estimate) is
    
    \begin{equation}
        \hat{x}_{\text{mean}}(i,j,\lambda) \;=\; \frac{1}{N}\sum_{k=1}^{N} x_0^{(k)}(i,j,\lambda).
    \end{equation}
    
    The \emph{uncertainty cube} is the per-pixel, per-band variance:
    \begin{equation}
        U(i,j,\lambda) \;=\; \operatorname{Var}_{k}\!\bigl[x_0^{(k)}(i,j,\lambda)\bigr].
    \end{equation}
   
\subsection{Training Data and Metameric Augmentation}

    Diffusion models are highly data-intensive, as they must learn to approximate the entire, high-dimensional manifold of a data distribution $p(\mathbf{x})$. Standard HSI datasets, while high-quality, are orders of magnitude smaller than typical image datasets used for training such models. Furthermore, as noted by \cite{fu_limitations_2025}, they are particularly deficient in metameric diversity. Training a diffusion prior $D_\theta$ on this limited distribution risks overfitting and failing to capture the true, ambiguous nature of the HSI-to-RGB mapping. To mitigate this and enrich the distribution's diversity, we conduct a comprehensive metameric augmentation strategy with two components.
    
    First, the distribution is enriched with ``black metamers'', i.e., spectral distributions $\mathbf{X}_{\text{black}}$ that integrate to zero under the SRF ($\mathbf{X}_{\text{black}} \textbf{Q} = 0$) 
    In~\cite{fu_limitations_2025}, metamers are generated by multiplying a random scalar to the black part. However, this leads to the same black metamer for the whole hyperspectral image. We introduce a novel, texture-guided approach where each training image is pre-segmented using the Segment Anything Model (SAM)~\cite{ravi_sam_2024} to identify distinct texture regions. The black metamer is then added with a region-wise factor $\alpha_r$, assigned uniformly within each texture segment to ensure spectral consistency, yet drawn independently from segment to segment. 
    This teaches the model that different physical textures can possess different degrees of spectral ambiguity. Full implementation details are provided in the Supplementary Material. Second, we adapt the Partition of Unity (PU) basis functions developed by~\cite{belcour_one--many_2023} to generate metamers for a given target RGB image. While the original work was formulated to generate metamers for a target in the CIE XYZ color space, we reformulate the basis functions to be specific to the sensor's RGB space. This adaptation allows us to generate a set of distinct spectra $\{\mathbf{x}_0, \mathbf{x}_0', \mathbf{x}_0'', ...\}$ that all map to exactly the same RGB measurement (i.e., $\mathcal{A}(\mathbf{x}_0) = \mathcal{A}(\mathbf{x}_0') = y$). By training the denoiser $D_\theta$ on these explicit metamer sets, the prior learns to recognize part of the null-space of the forward operator, which is critical for correctly modeling the posterior variance during inference. Combining these two metamer generation methods, we significantly diversify existing datasets and expose the network with more metamer cases.

\section{Experiments}
\label{sec:experiments}

We train the unconditional diffusion prior $D_\theta$ on ARAD-1K \cite{arad_ntire_2022} and KAUST \cite{li_multispectral_2021}, yielding $\sim 1400$ high-resolution HSIs spanning indoor/outdoor scenes and diverse textures. Validation is performed on ICVL \cite{arad_sparse_2016} and CAVE \cite{yasuma_generalized_2010}. All cubes are resampled to 31 bands on $[400,700]$ nm, values normalized to $[-1,1]$, and randomly cropped to $256\times256$. For posterior inference we evaluate five forward operators $\mathcal{A}$ that differ in their wavelength-dependent PSF: (1) None (naïve pixel-wise SRF mixing), (2) Grating PSF~\cite{baek_compact_2017}, (3) Gaussian PSF (chromatic aberration induced by a Double Gaussian lens~\cite{ichimura2021optical}), (4) Rotational diffraction PSF (a DOE that imparts wavelength-dependent rotation)~\cite{jeon_compact_2019}, and (5) CASSI~\cite{meng2020end}. We use stochastic guided EDM sampling~\cite{karras_elucidating_2022, chakraborty_multimodal_2025} with $20–40$ NFEs, and $N=20$ posterior samples per image.

\begin{figure}
    \centering
    \includegraphics[width=\linewidth]{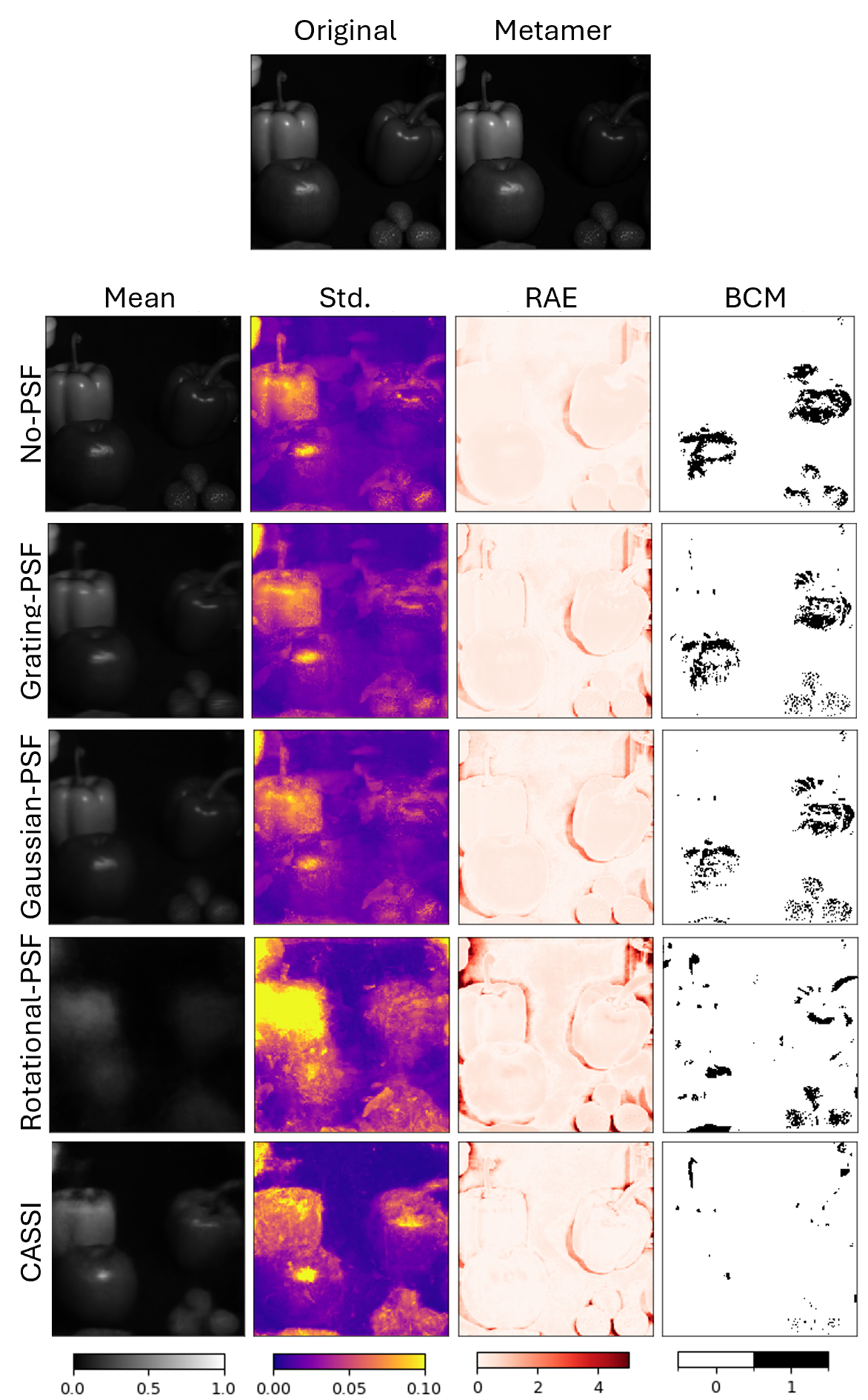}
    \caption{HSDiff results on CAVE. Top: original and metamer HSI cubes, averaged over the spectral dimension. Each subsequent row corresponds to the operator used in guided diffusion—None (no aberration), Grating PSF, Gaussian PSF, Rotational-diffraction PSF, and CASSI. Columns show: (i) posterior mean from 20 samples, (ii) posterior standard deviation, (iii) absolute error $|\text{mean} - \text{original}|$, and (iv) Binary Coverage Map (BCM), where black indicates ground truth outside the $95\%$ prediction interval.
}
\vspace{-5mm}
    \label{fig:grid}
\end{figure}

Figure \ref{fig:grid} contrasts operators within our guided posterior sampling on CAVE. The results are averaged over wavelength to visualize the scene layout. For each operator (None, Grating, Gaussian, Rotational-diffraction, CASSI), we report four summaries: the posterior mean from 20 samples, the standard deviation (uncertainty), the absolute error $|\text{mean}-\text{original}|$, and a Binary Coverage Map (BCM) marking pixels where the ground truth falls outside the nominal $95\%$ prediction interval. Consistent with the spatial analyses, None, Gaussian, and Grating recover finer structure with lower error, while the Rotational PSF produces blurrier details and slightly higher variance. In terms of confidence calibration, Rotational and CASSI achieve the highest coverage in the BCM, while other approaches are more likely to be ``confidently wrong''.

\begin{figure*}
    \centering
    \includegraphics[width=\linewidth]{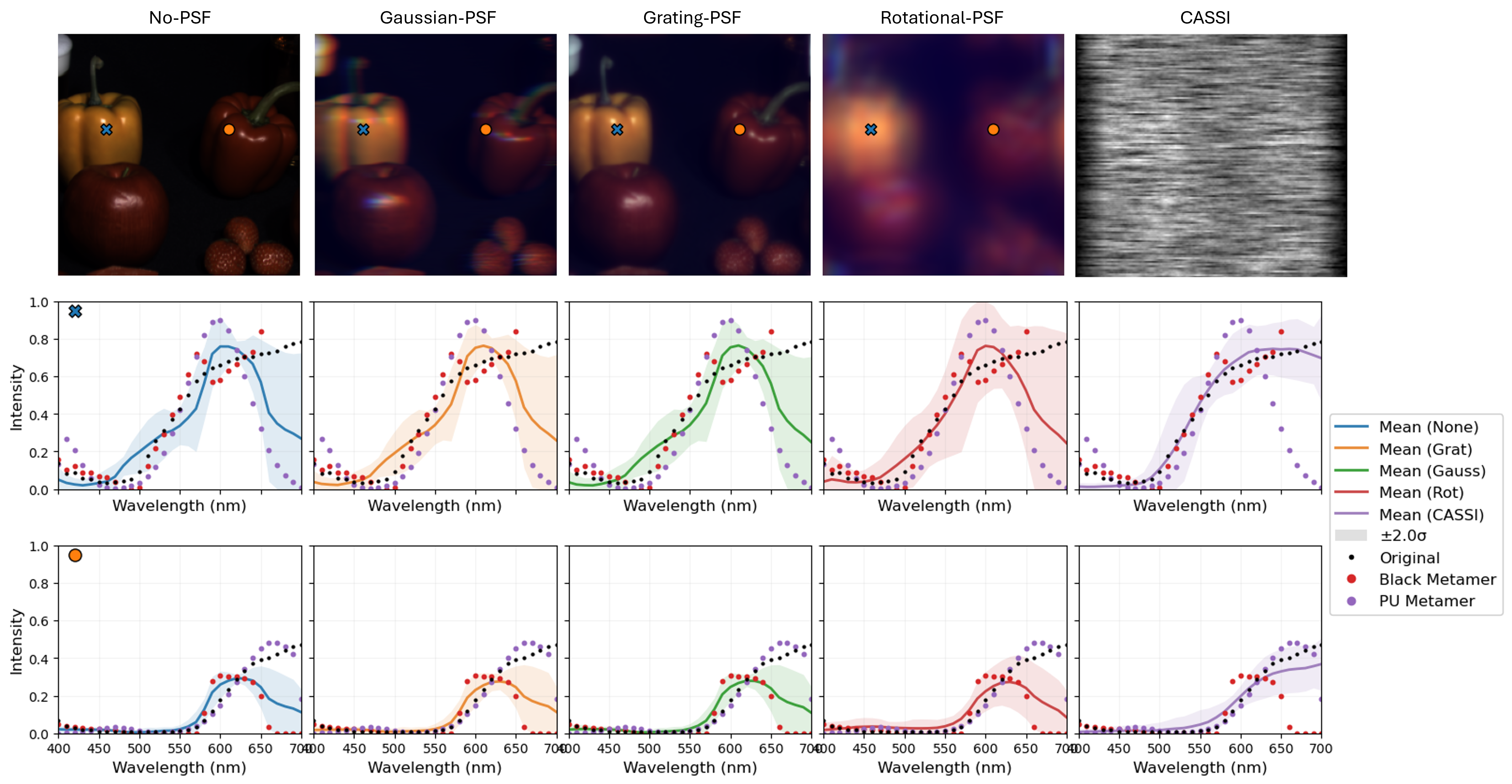}
    \caption{Top: RGB images under no aberration, Gaussian PSF, Grating PSF, Rotational PSF, and the CASSI measurement. Middle and bottom: spectral profiles at the marked pixels. Each plot shows the original spectrum, one sample black metamer, and one PU-basis metamer, along with the posterior mean $\pm \ 2\sigma$ after guided diffusion. The metamer-enriched prior broadens the credible bands—posterior samples typically lie between the ground truth and the metameric alternatives. In contrast, CASSI remains close to the ground truth, since its measurement is not affected by RGB-targeted metamerism.}
    \label{fig:profiles}
\end{figure*}

Figure \ref{fig:profiles} shows spectral uncertainty at two pixels (the yellow and red peppers) for both the original and the metamer data. For each pixel, we plot the ground-truth spectrum, two sample metamers (black metamer and PU-basis metamer), and the posterior summaries produced by each operator (PSF variants and CASSI). Two consistent patterns emerge. First, for optics-aware operators (PSFs and CASSI), the posterior mean typically falls between the original spectrum and the corresponding metamer, indicating that the sampler does not collapse to a single spectrum but instead reflects the metameric ambiguity present in RGB measurements. Second, the credible bands (shaded mean $\pm \ 2 \sigma$) expand in wavelength regions where the forward operator is less discriminative and contract where the operator supplies stronger spectral cues—indicating informative uncertainty.

Because the diffusion prior was trained with metameric augmentations (black and PU-basis), the posterior samples are sufficiently expressive to represent both families of spectra, which improves robustness of uncertainty quantification: variance rises precisely where the metamer and original diverge, and decreases when the operator breaks the ambiguity. CASSI’s credible bands concentrate around the original spectrum because RGB metamers do not remain indistinguishable under CASSI’s spectral encoding—each metamer maps to a different CASSI measurement. As a result, the posterior is better constrained and the uncertainty narrows toward the true spectrum, unlike in RGB-only settings where metamerism broadens the credible bands. Additional examples are provided in the Supplementary Material.

\begin{figure}[!h]
    \centering
    \includegraphics[width=\linewidth]{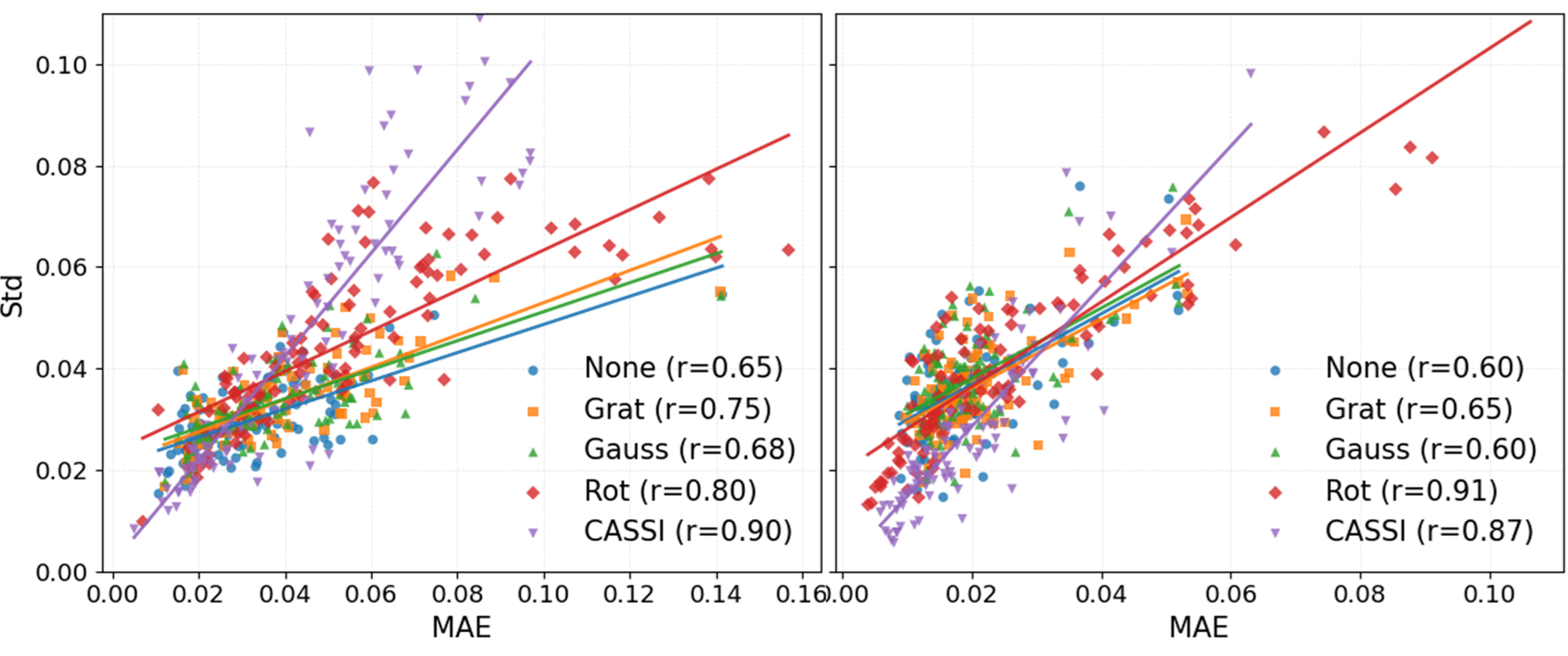}
    \caption{Calibration cross-plots (mean std vs MAE). Left: CAVE. Right: ICVL. Each point is one image–operator pair (None, Grating, Gaussian, Rotational, CASSI). A clear positive trend indicates informative uncertainty.}
    \label{fig:crossplot}
\end{figure}

To assess the practical value of the predicted uncertainties, Figure \ref{fig:crossplot} analyzes calibration on CAVE and ICVL. For each validation image and each operator (PSF variants and CASSI), we plot a point with the mean absolute error (MAE) of the posterior mean w.r.t. the original HSI, aggregated over all pixels and bands, and the corresponding posterior standard deviation, aggregated over the same domain. Each point, therefore, summarizes one image. A clear positive trend indicates that higher reconstruction error is accompanied by higher predicted uncertainty, i.e., the uncertainty is informative. Across methods, all show a positive correlation; Rotational and CASSI exhibit the strongest correlation, meaning they most reliably flag hard (high-error) images with larger uncertainty. An ablation study for the diffusion sampling hyperparameters is shown in the Supplementary Material.

\begin{figure}
    \centering
    \includegraphics[width=\linewidth]{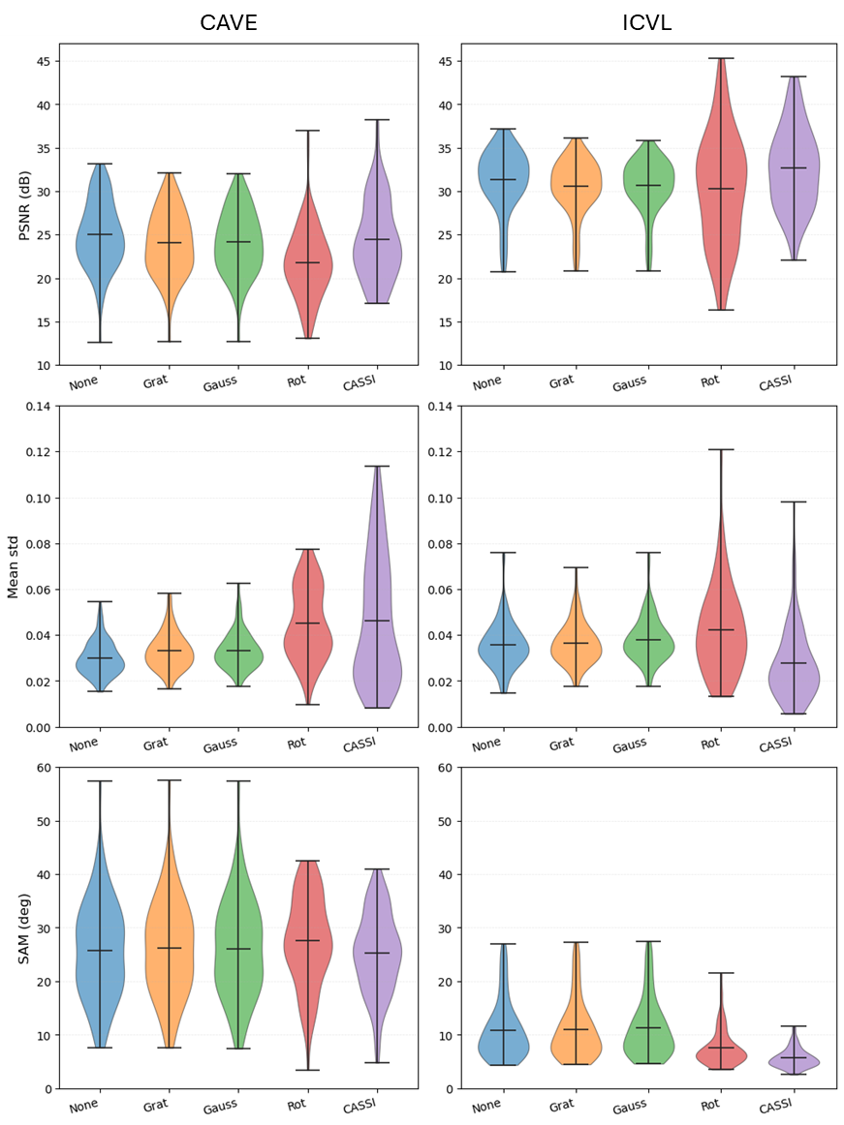}
    \caption{Per-image metrics across operators. Violin plots for CAVE (left) and ICVL (right): PSNR (top, higher is better), mean posterior std (middle, uncertainty proxy), and SAM (bottom, lower is better). CASSI trends higher PSNR/lower SAM overall.}
    \label{fig:boxplot}
\end{figure}

A summary of per-image metrics by operator for CAVE (left) and ICVL (right) is shown in Figure ~\ref{fig:boxplot}. On ICVL, CASSI consistently achieves the highest PSNR and lowest SAM, while also exhibiting the lowest mean std, indicating accurate and confident reconstructions; Rotational is second in PSNR but carries noticeably higher uncertainty. On CAVE, CASSI again delivers the best (lowest) SAM with competitive PSNR, yet its mean std is higher. In contrast, Rotational underperforms on both PSNR and SAM and shows elevated uncertainty, whereas Grating and Gaussian occupy a middle ground with tighter spreads across images. Overall, it is shown that optics-aware encoding improves fidelity, and the uncertainty estimates track task difficulty instead of remaining uniformly low.

A likely reason for the weaker performance on CAVE relative to ICVL is domain shift. Although CAVE is a small, laboratory-style dataset with many scenes dominated by large black backgrounds, controlled illumination, but diverse materials (real and fake); these statistics differ markedly from our training distributions in ARAD1K and KAUST (natural/complex scenes with richer textures and albedos). In contrast, ICVL offers broader content and illumination variability that better matches the training data, leading to more reliable likelihood gradients and tighter posteriors. 

Tables \ref{tab:metrics-cave} and \ref{tab:metrics-icvl} rank the methods by reconstruction quality of the sample mean and their ability to produce calibrated uncertainty estimates. Across both datasets, the CASSI operator delivers the best reconstruction quality, achieving the highest PSNR and lowest SAM. It also demonstrates strong error-uncertainty coupling (high Pearson correlation) and solid calibration, characterized by high coverage (PICP) with a low posterior std. The Rotational PSF lags in fidelity and exhibits higher uncertainty. Despite this, it provides more conservative coverage (higher PICP) and maintains a strong correlation between error and uncertainty. The remaining methods (None, Grating, Gauss) perform in the mid-range for fidelity. Their calibration behavior is inconsistent across datasets: on CAVE, the None operator appears over-confident (lowest std, weaker correlation), whereas on ICVL, these baselines attain high coverage but do not match the fidelity of CASSI. Overall, the results show that, while most optics-aware encodings improve fidelity over the non-encoded baseline, the specific operator is critical. The calibration behavior varies systematically with the operator, and CASSI provides the best balance of high-fidelity reconstruction and well-calibrated uncertainty.

\begin{table}[t]
    \centering
    \caption{Reconstruction quality and uncertainty calibration by operator for the CAVE dataset.}
    \label{tab:metrics-cave}
    \footnotesize
    \begin{tabular}{l RR UUU}
    \toprule
    & \multicolumn{2}{c}{\shortstack{\textbf{Reconstruction}\\\textbf{quality}}}
    & \multicolumn{3}{c}{\shortstack{\textbf{Uncertainty \&}\\\textbf{calibration}}} \\
    \cmidrule(lr){2-3} \cmidrule(lr){4-6}
    Method & PSNR $\uparrow$ & SAM $\downarrow$ & Pearson $\uparrow$ & PICP $\uparrow$ & Std $\downarrow$ \\
    \midrule
    None  & \second{22.07} & \second{25.41} & 0.65          & 0.8137 & \textbf{0.032} \\
    Grating  & 21.67          & 25.67          & 0.75          & 0.8134 & \second{0.034} \\
    Gaussian & 21.66          & 25.66          & 0.68          & 0.8162 & 0.035 \\
    Rotational  & 19.79    & 27.66          & \second{0.80} & \textbf{0.8258} & 0.048 \\
    CASSI & \textbf{22.81} & \textbf{23.85} & \textbf{0.90} & \second{0.8213} & 0.045 \\
    \bottomrule
    \end{tabular}
\end{table}

\begin{table}[t]
    \centering
    \caption{Reconstruction quality and uncertainty calibration by operator for the ICVL dataset.}
    \label{tab:metrics-icvl}
    \footnotesize
    \begin{tabular}{l RR UUU}
    \toprule
    & \multicolumn{2}{c}{\shortstack{\textbf{Reconstruction}\\\textbf{quality}}}
    & \multicolumn{3}{c}{\shortstack{\textbf{Uncertainty \&}\\\textbf{calibration}}} \\
    \cmidrule(lr){2-3} \cmidrule(lr){4-6}
    Method & PSNR $\uparrow$ & SAM $\downarrow$ & Pearson $\uparrow$ & PICP $\uparrow$ & Std $\downarrow$ \\
    \midrule
    None  & \second{30.52} & 10.01          & 0.60          & \textbf{0.9891} & \second{0.035} \\
    Grating  & 30.03          & 10.17          & 0.65          & 0.9864          & 0.036 \\
    Gaussian & 30.05          & 10.50          & 0.60          & \second{0.9888} & 0.037 \\
    Rotational  & 27.89    & \second{7.43}  & \textbf{0.91} & 0.9720          & 0.039 \\
    CASSI & \textbf{31.14} & \textbf{5.87}  & \second{0.87} & 0.9601          & \textbf{0.029} \\
    \bottomrule
    \end{tabular}
\end{table}

\subsection{Limitations}
    Diffusion priors are data-hungry: despite metamer-aware augmentation (black and PU-basis metamers) and optics modeling, our prior is ultimately constrained by the diversity of the training set. Small or biased corpora (limited materials, illuminants, camera SRFs/CSSs, and PSF variants) can induce domain shift. Expanding dataset breadth (materials/illuminations), improving forward-model estimation (for different noise types), and optimizing more effective spectral encoding are key to mitigating these limitations.
\section{Conclusions}
\label{sec:conclusions}

Despite seemingly high scores being reported for hyperspectral image reconstruction in the literature, their deterministic inference may result in over-confident failures in challenging scenes, such as metamers. Our proposed Bayesian framework quantifies the uncertainty with a diffusion model where pixel-level diffusion priors are learned from metamer-augmented data. In contrast to existing scalar metameric black augmentation, our enhanced metmaric spectral augmentation leverages texture-guided metamer generation with both black metamers and partition-of-union basis, substantially increasing the spectral diversity of current datasets. We conduct cross-dataset training and validation to quantitatively evaluate the performance of the framework. The results indicate that, with effective spectral encoding, our method achieves improved reconstruction accuracy together with calibrated predictive uncertainty. More importantly, the findings provide compelling evidence that effective spectral encoding is essential for mitigating metameric ambiguities. Further performance gains, however, will require more deliberate optimization of the spectral encoding scheme beyond existing options analyzed in this work.
\clearpage
{
    \small
    \bibliographystyle{ieeenat_fullname}
    \bibliography{main}
}



\clearpage
\setcounter{page}{1}
\maketitlesupplementary

\renewcommand{\thefigure}{S\arabic{figure}}
\renewcommand{\thetable}{S.\Roman{table}}
\renewcommand{\theequation}{S\arabic{equation}}

\appendix

\section{Spectral Encoding and Probabilistic Modeling}
\label{sec:nll}

As discussed in the Introduction, many state-of-the-art RGB $\rightarrow$ HSI models report metrics on only one or two small benchmarks, yet behave like overfit regressors. They learn an almost deterministic mapping that does not transfer across datasets or camera/operators. This fragility stems from two factors: (i) the inherent ill-posedness of mapping 1 or 3 channels to $K=31$ bands, and (ii) limited, biased training data. As shown by \cite{fu_limitations_2025}, introducing optics—aware spectral encoding, such as wavelength-dependent PSFs, can improve identifiability by injecting spatially varying, band-specific cues. In this section, we further illustrate that incorporating uncertainty during training exposes how deterministic models can be confidently wrong for metameric cases.

\paragraph{From deterministic to probabilistic supervision.}
    To probe these effects without changing architectures, we first convert a strong deterministic baseline to output Gaussian predictions via an NLL loss (details in the next section). This produces per-pixel, per-band means and (diagonal) variances and lets us study how spectral encoding (e.g., optical aberrations) affects the posterior spread of the inverse problem. Figure 1a in the main paper illustrates the limitation of this approach: although supervised NLL captures uncertainty, metamerism persists, i.e., multiple spectra produce the same RGB, so genuine metamer spectra often lie \textbf{outside the learned confidence bands}. In other words, a diagonal-Gaussian predictor trained on a single ground-truth target tends to underestimate epistemic ambiguity and cannot represent multi-modal posteriors induced by metamers.

\paragraph{Effect of optical encoding and metamer exposure.}
    We then introduce wavelength-dependent blur (PSFs) and explicitly expose the network to one black metamer for every input image during training. As shown in Figure 1b, optical encoding helps separate otherwise confounded spectra: the same RGB and its metamer yield different PSF-blurred images, reducing the predicted intervals around each respective spectra. This preliminary experiment confirms the findings in \cite{fu_limitations_2025}: optical encoding (via the PSF) provides spatial cues that constrain the ill-posed problem, and metamer augmentation is key for a realistic and diverse posterior distribution. 

    \paragraph{Requirements for a Robust HSI Framework.} A framework for reliable and calibrated uncertainty-aware HSI reconstruction must satisfy three core requirements:
    
    \begin{enumerate}
        \item Data Diversity: It must leverage substantially broader data diversity (materials, illuminants, SRFs, operators) than is available in standard datasets, which can be achieved through new acquisition, synthetic generation, or rigorous spectral augmentation.
        \item Operator Flexibility: It must support arbitrary forward models (e.g., different PSFs) to inject discriminative spectral cues without requiring a full retraining for each new operator.
        \item Posterior Expressivity: It must model complex, non-Gaussian posteriors that capture the spatio-spectral correlations and potential multimodality inherent in HSI data.
    \end{enumerate}
    
    A supervised network (e.g., MST++ with NLL training), unfortunately, fails to meet these requirements. While it can be trained with augmented data for a specific set of PSFs (partially addressing Req. 1), it fails on the other two. It lacks expressivity (failing Req. 3), as its NLL objective typically assumes an oversimplified, independent Gaussian posterior. It also lacks flexibility (failing Req. 2): the operator is ``baked in'' to the network weights, and a new PSF would require a complete retraining.
    
    In contrast, off-the-shelf diffusion priors when trained on an augmented distribution, learns a rich, diverse data model (meeting Req. 1 \& 3). When paired with guided sampling (e.g., DPS), it can natively incorporate any arbitrary, differentiable operator at inference time. This combination simultaneously meets all three requirements and, critically, without retraining the prior, enables the systematic evaluation of how different forward models reshape the posterior solution space.
    This synthesis that combines optics-aware physics and posterior sampling with diffusion priors is the core motivation behind our HSDiff framework.

\subsection{Probabilistic Benchmarking}
    
    Unlike deterministic evaluation, a fair probabilistic comparison must assess three aspects jointly: (i) accuracy -- how close the posterior mean is to the ground truth (e.g., PSNR, SAM); (ii) uncertainty magnitude -- how large the posterior spread is (e.g., mean/std over pixels and bands, interval width); and (iii) calibration/informativeness -- whether larger predicted uncertainty coincides with larger errors, and if the ground truth is contained in the confidence interval.
    
    A second prerequisite is comparability of the uncertainty itself. Many probabilistic deep-learning approaches (Bayesian neural nets, ensembles, heteroscedastic regressors) yield parameter-conditional uncertainty, i.e., variability induced by weight/posterior approximations, whereas our target is the data-conditional posterior over HSIs given the same forward model and noise. To avoid mixing fundamentally different uncertainty notions, all methods must be evaluated under an identical likelihood (same forward operator and noise model) and report uncertainty over the same variables (per-pixel, per-band spectra, aggregated as specified).
    
    Ideally, benchmarking HSDiff would involve other deep priors with full Bayesian inference. While some frameworks do exist, public code/weights are scarce. As a pragmatic baseline, we therefore convert a strong deterministic architecture to a heteroscedastic Gaussian predictor trained with NLL, yielding per-band diagonal covariances. This baseline lets us: (a) introduce probabilistic evaluation on familiar models; (b) test the impact of spectral encoding and metameric augmentation on uncertainty; and (c) compare against HSDiff under the same forward model, noise, datasets, and splits. 
    
\begin{figure*}[ht]
    \centering
    \includegraphics[width=0.6\linewidth]{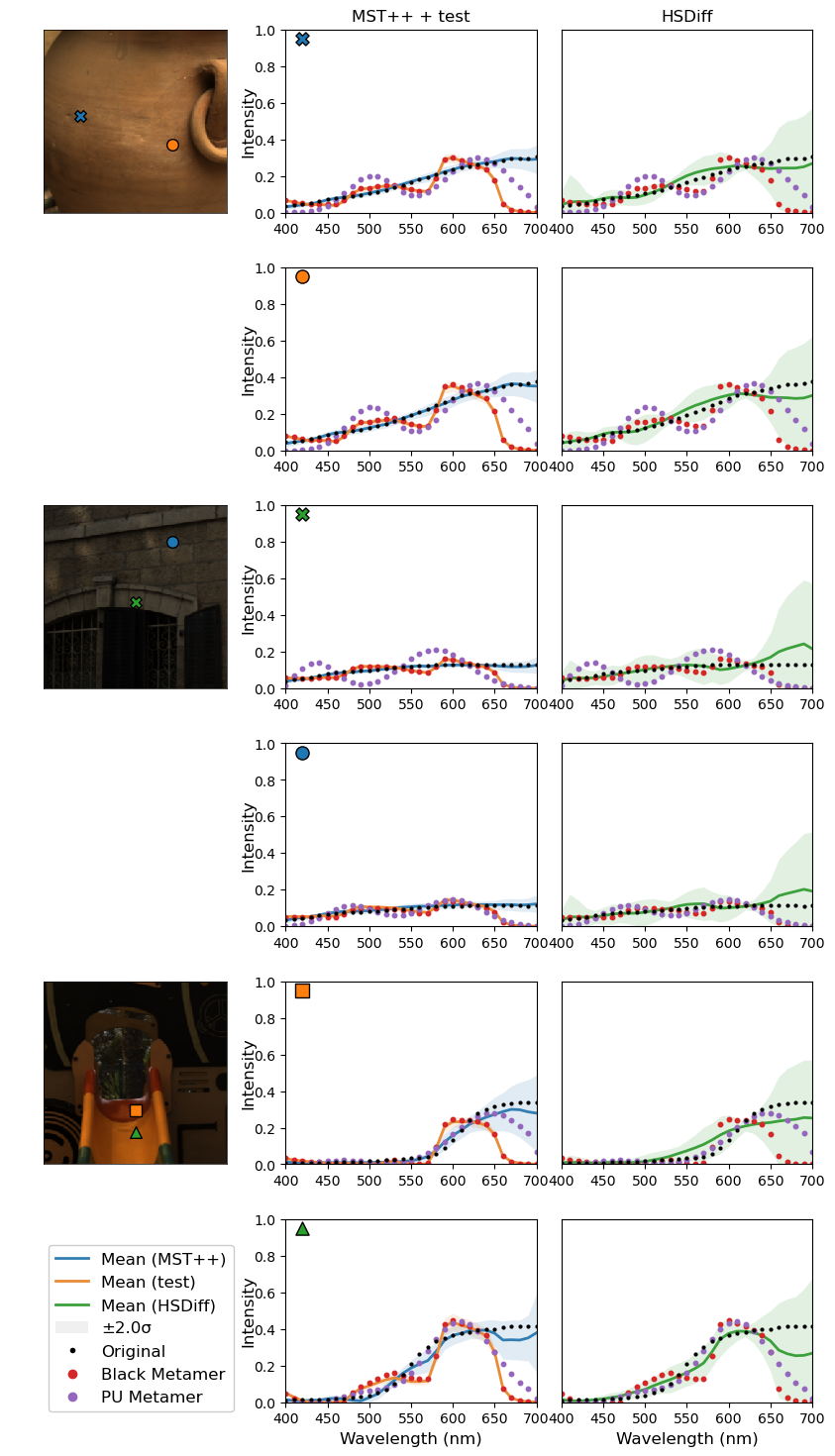}
    \caption{MST++ (NLL) vs. HSDiff under Gaussian PSF. MST++ is trained with doubled inputs (original + black-metamer) and Gaussian-PSF RGB targets, leading to posterior bands concentrated around seen spectra (original/black). HSDiff produces broader, operator-aware credible bands that better cover both black and PU-basis metamers, indicating more calibrated uncertainty. Note that calibration remains bounded by training diversity (illumination, materials, camera SRFs) and the set of metamers encountered.
}
    \label{fig:mstvshsdiff}
\end{figure*}

    Figure \ref{fig:mstvshsdiff} compares MST++ (NLL) with HSDiff. For MST++, we doubled the training inputs by pairing each original spectrum with a black-metamer variant, and synthesized the targets as Gaussian-PSF RGB measurements. As a result, the MST++–NLL posterior concentrates around the spectra it has seen during training (original and black metamer), yielding comparatively narrow credible bands in Figure \ref{fig:mstvshsdiff}. However, because only one metamer family was injected (from an infinite set), other valid explanations, e.g., the PU-basis metamer, often fall outside these intervals. In contrast, HSDiff produces broader, operator-aware credible bands that better cover both the black and PU-basis metamers, indicating more calibrated uncertainty over the space of metameric solutions. However, this should not be interpreted as an “ultimate” solution: even with PSF guidance and metamer augmentation, coverage remains bounded by the diversity of training conditions (illumination, materials, camera SRFs) and the set of metamers encountered. Expanding spectral/scene diversity, as well as operator variation, would further improve calibration.
    
    Note that HSDiff posterior $p(\mathbf{x}\mid\mathbf{y})$ is obtained from an explicit prior--likelihood factorization $p(\mathbf{y}\mid\mathbf{x})\,p(\mathbf{x})$, whereas an NLL-trained network learns a parametric conditional $p_\theta(\mathbf{x}\mid\mathbf{y}) = p(\mathbf{x}\mid\mathbf{y},\theta)$ directly from data, without an explicit separation between prior and measurement model. As a result, the associated uncertainty estimates have different weights: in the former, credible regions reflect the Bayesian combination of a generative prior and a physically motivated likelihood, while in the latter, variances are learned surrogates shaped by the model class, parameterization, and training data. Directly comparing the numerical scale of these two uncertainties can therefore be direct. 

\section{NLL Training}
Consider a supervised learning setting with inputs $\mathbf{y}\in\mathbb{R}^m$ (e.g., measurements) and targets $\mathbf{x}\in\mathbb{R}^n$ (e.g., reconstructions). A standard deterministic neural network implements a mapping
\begin{equation}
    f_\theta : \mathbf{y} \mapsto \hat{\mathbf{x}} = f_\theta(\mathbf{y}),
\end{equation}
with parameters $\theta$, and is typically trained by minimizing a pointwise loss such as the mean squared error (MSE),
\begin{equation}
    \mathcal{L}_{\text{MSE}}(\theta)
    = \frac{1}{N}\sum_{i=1}^N
    \bigl\| \mathbf{x}_i - f_\theta(\mathbf{y}_i) \bigr\|_2^2.
\end{equation}

To obtain a probabilistic model, we interpret the network output as the parameters of a conditional distribution $p_\theta(\mathbf{x}\mid \mathbf{y})$ and train by minimizing the negative log-likelihood (NLL). A common choice is a Gaussian likelihood
\begin{equation}
    p_\theta(\mathbf{x}\mid \mathbf{y})
    = \mathcal{N}\bigl(
        \mathbf{x} \mid
        \boldsymbol{\mu}_\theta(\mathbf{y}),\,
        \Sigma_\theta(\mathbf{y})
    \bigr),
\end{equation}
where $\boldsymbol{\mu}_\theta(\mathbf{y})\in\mathbb{R}^n$ is the predicted mean and $\Sigma_\theta(\mathbf{y})\in\mathbb{R}^{n\times n}$ the predicted covariance. In practice, we often use a diagonal covariance,
\begin{equation}
    \Sigma_\theta(\mathbf{y})
= \operatorname{diag}\bigl(\boldsymbol{\sigma}^2_\theta(\mathbf{y})\bigr),
\end{equation}
and let the network output both $\boldsymbol{\mu}_\theta(\mathbf{y})$ and (for numerical stability) $\log \boldsymbol{\sigma}^2_\theta(\mathbf{y})$. Given a dataset $\{(\mathbf{y}_i,\mathbf{x}_i)\}_{i=1}^N$, the NLL loss for the diagonal Gaussian case is
\begin{align}
    \mathcal{L}_{\text{NLL}}(\theta)
    &= -\frac{1}{N}\sum_{i=1}^N
       \log p_\theta(\mathbf{x}_i\mid \mathbf{y}_i) \\
    &= \frac{1}{2N}\sum_{i=1}^N
       \left[
           \frac{\bigl\|\mathbf{x}_i - \boldsymbol{\mu}_\theta(\mathbf{y}_i)\bigr\|_2^2}
                {\boldsymbol{\sigma}_\theta^2(\mathbf{y}_i)}
           + \log \boldsymbol{\sigma}_\theta^2(\mathbf{y}_i)
       \right].
\end{align}

Minimizing $\mathcal{L}_{\text{NLL}}$ trains $f_\theta$ as a probabilistic predictor: for each input $\mathbf{y}$, the network outputs a full conditional distribution $p_\theta(\mathbf{x}\mid\mathbf{y})$, providing both a point estimate (through $\boldsymbol{\mu}_\theta$) and an uncertainty estimate (through $\boldsymbol{\sigma}_\theta$). In this work, we employ this approach to transform a state-of-the-art deterministic neural network architecture for RGB to HSI MST++ \cite{cai_mst_2022}, enabling the prediction of probabilistic estimates. A sample of those results is shown in Figure 1 in the main paper.

\section{Guided Diffusion Sampling in the EDM framework}
\label{sec:sup_edm}
In Equation (11) in the main paper, we introduce posterior sampling using a pre-trained diffusion prior and a likelihood score. The likelihood term can be written as \cite{chung_diffusion_2024}
\begin{equation}
    \label{eqn:likelihood_}
    p(\mathbf{y}\mid \mathbf{x}_t)
    = \int p(\mathbf{y}\mid \mathbf{x}_0)\, p(\mathbf{x}_0\mid \mathbf{x}_t)\,\mathrm{d}\mathbf{x},
\end{equation}
where $\mathbf{x}_t \in \mathbb{R}^n$ denotes the noisy latent at time $t$ and $\mathbf{x}\in\mathbb{R}^n$ the clean latent. Assuming Gaussian observation noise,
\begin{equation}
    p(\mathbf{y}\mid \mathbf{x}_0) = \mathcal{N}\bigl(\mathbf{y} \mid A \mathbf{x}_0,\, \Sigma_y\bigr),
\end{equation}
where $A\in\mathbb{R}^{m\times n}$ is the linear forward operator and $\Sigma_y\in\mathbb{R}^{m\times m}$ is the noise covariance. Following \cite{rozet_score-based_2023}, the conditional over the clean latent is approximated as
\begin{equation}
    p(\mathbf{x}_0\mid \mathbf{x}_t) \approx
    \mathcal{N}\bigl(\mathbf{x} \mid \hat{\mathbf{x}}_0(\mathbf{x}_t),\, \Sigma_{\mathbf{x}\mid t}\bigr),
\end{equation}
where $\hat{\mathbf{x}}_0(\mathbf{x}_t)$ is the posterior mean and $\Sigma_{\mathbf{x}_0\mid t}$ its covariance, approximated by
\begin{equation}
    \Sigma_{\mathbf{x}\mid t} \approx \frac{\sigma(t)^2}{\mu(t)^2}\,\Gamma,
\end{equation}
with $\Gamma$ a fixed positive-definite matrix that summarizes the prior covariance \cite{rozet_score-based_2023}. Here $\mu(t)$ and $\sigma(t)$ are the scalar mean and standard-deviation schedules of the linear Gaussian forward process which corrupts the clean signal $\mathbf{x}_0$ into $\mathbf{x}_t$. In the case $A$ is linear, the integral in Eq.~\eqref{eqn:likelihood_} is an exact Gaussian marginalization. Writing $\mathbf{x}_0 = \hat{\mathbf{x}}_0(\mathbf{x}_t) + \boldsymbol{\varepsilon}_x,
 $ with $\boldsymbol{\varepsilon}_x\sim\mathcal{N}(0,\Sigma_{\mathbf{x}_0\mid t})$ and $
\mathbf{y} = A \mathbf{x} + \boldsymbol{\varepsilon}_y$ with $ \boldsymbol{\varepsilon}_y\sim\mathcal{N}(0,\Sigma_y)$ we obtain:
\begin{align}
    \mathbb{E}[\mathbf{y} \mid \mathbf{x}_t] &= A\,\hat{\mathbf{x}}_0(\mathbf{x}_t), \\
    \operatorname{Cov}(\mathbf{y} \mid \mathbf{x}_t) &= A\,\Sigma_{\mathbf{x}_0\mid t}\,A^\top + \Sigma_y.
\end{align}
Substituting the approximation for $\Sigma_{\mathbf{x}\mid t}$ yields the perturbed likelihood
\begin{equation}
    p(\mathbf{y}\mid \mathbf{x}_t)
    \approx
    \mathcal{N}\!\left(
        \mathbf{y} \,\middle|\,
        A\hat{\mathbf{x}}_0(\mathbf{x}_t),\,
        \Sigma_y + \frac{\sigma(t)^2}{\mu(t)^2}A\Gamma A^\top
    \right).
\end{equation}

This likelihood was introduced in Eq.~\eqref{eqn:likelihood_} for a general (possibly non-linear) operator $A$. In our EDM setting, we use the parameterization with unit signal scaling, so $\mu(t)=1$ and therefore $\sigma(t)^2/\mu(t)^2 = \sigma(t)^2$. Moreover, we approximate $A\Gamma A^\top \approx \nu I$ and $\Sigma_y \approx \sigma_y I$, with $\nu,\sigma_y > 0$ scalars, which yields
\begin{equation}
    p(\mathbf{y}\mid \mathbf{x}_t)
    \approx
    \mathcal{N}\!\left(
        y \,\middle|\,
        A\hat{\mathbf{x}}_0(\mathbf{x}_t),\,
        \bigl(\sigma_y + \sigma(t)^2 \nu\bigr) I
    \right).
\end{equation}                                      

For the diffusion sampling, we use the stochastic EDM sampler \cite{karras_elucidating_2022}, adapted to the previous perturbed likelihood formalism \cite{chakraborty_multimodal_2025}. The algorithm is shown in Algorithm \ref{alg:guided_edm}.

\section{Metameric Augmentation}
\label{sec:sup_metamers}

\subsection{Texture guided black-metamers}
\begin{figure*}[ht]
    \centering
    \includegraphics[width=\linewidth]{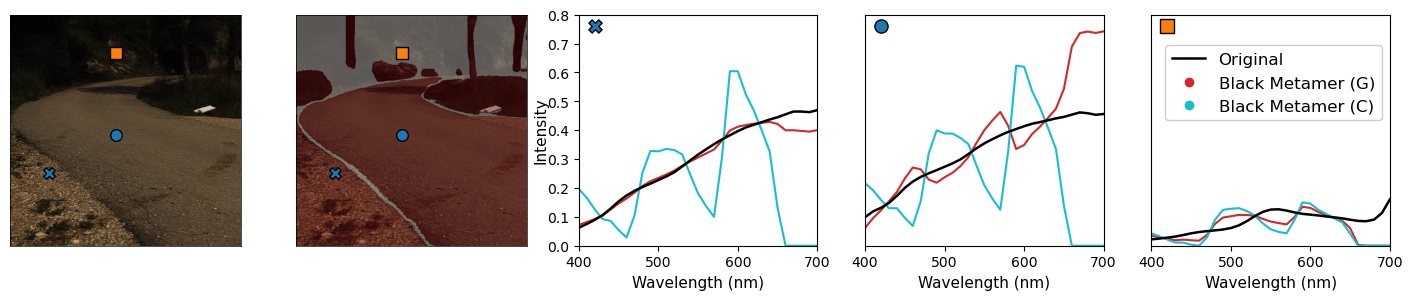}
    \caption{Black metamer generation (constant (C) vs. texture-guided (G) $\alpha$). For the 3 pixel locations indicated in the left RGB and mask images, a constant $\alpha$ produces metamers with similar geometry across materials. On the other hand, texture-guided $\alpha$ yields material-dependent metamer geometry, enabling texture-consistent augmentation and greater spectral diversity.
}
    \label{fig:met_comp}
\end{figure*}

    We generate camera-metamers through a fundamental and metameric black decomposition. Let $\mathbf{S} \in \mathbb{R}^{K}$ denote a per-pixel spectrum and $\mathbf{Q} \in \mathbb{R}^{K \times 3}$ the camera spectral responses. The orthogonal projector onto the sensor subspace $\mathrm{col}(\mathbf{Q})$ is
    \begin{equation}
        \mathbf{R} \;=\; \mathbf{Q}\,(\mathbf{Q}^\top \mathbf{Q})^{-1}\mathbf{Q}^\top.
    \end{equation}
    The fundamental metamer is $\mathbf{S}_0 = \mathbf{R}\mathbf{S}$; the metameric black (sensor-invisible component) is
    \begin{equation}
        \mathbf{S}_b \;=\; (\mathbf{I} - \mathbf{R})\,\mathbf{S} \;\in\; \mathrm{null}(\mathbf{Q}^\top).
    \end{equation}
    Any spectrum of the form
    \begin{equation}
        \mathbf{S}_x(\alpha) \;=\; \mathbf{S}_0 \;+\; \alpha\, \mathbf{S}_b
    \end{equation}
    produces the same camera response because $\mathbf{Q}^\top \mathbf{S}_b = \mathbf{0}$, hence $\mathbf{Q}^\top \mathbf{S}_x = \mathbf{Q}^\top \mathbf{S}$. More generally, with a nullspace basis $\mathbf{N} \in \mathbb{R}^{K \times d}$ ($\mathbf{Q}^\top \mathbf{N} = \mathbf{0}$), one can sample $\mathbf{S}_x = \mathbf{S}_0 + \mathbf{N}\mathbf{z}$ with $\mathbf{z} \in \mathbb{R}^{d}$ to explore the full metamer set. To promote a material-consistent $\alpha$, we employ a texture-guided approach. We first segment each RGB image using SAM2 \cite{ravi_sam_2024-1}, obtaining a mask with $N$ classes $\{\mathcal{C}_1,\dots, \mathcal{C}_N\}$. For every class $\mathcal{C}_k$ we draw a scalar
    \begin{equation}
        \alpha_k \sim \mathcal{U}(-1, 2),
    \end{equation}
    and use it for all pixels $\mathbf{S}$ belonging to that class:
    \begin{equation}
        \mathbf{S}'(\mathbf{p}) = \mathbf{S}^\ast(\mathbf{p}) +
        \alpha_{c(\mathbf{p})}\,\mathbf{B}(\mathbf{p}),
    \end{equation}
    where $c(\mathbf{p})\in\{1,\dots,N\}$ is the class index of pixel $\mathbf{p}$. This yields piecewise-constant metameric perturbations that respect object boundaries while generating diverse black metamers across the scene. We enforce spectra bounds by clipping values to $[0,1]$; see \cite{fu_limitations_2025} for a discussion of clipping effects. Figure \ref{fig:met_comp} compares black metamers generated with a constant $\alpha$ versus a texture-guided $\alpha$. Using a scalar $\alpha$, as in~\cite{fu_limitations_2025}, imposes the same spectral perturbation geometry across materials, whereas our texture guidance produces material-consistent variations. Different metamer geometries per region now yield a more diverse and realistic augmentation.

\subsection{PU-basis metamers}

    To further increase the diversity of metamers, we also employ a partition-of-union (PU) basis metamer generation as introduced in a recent paper~\cite{belcour_one--many_2023}. Let $\mathbf{S}\in\mathbb{R}^{K}$ denote a per-pixel spectrum on a $K$-sample wavelength grid and $\mathbf{Q}\in\mathbb{R}^{K\times 3}$ the camera spectral responses (camera RGB $\mathbf{t}=\mathbf{Q}^\top\mathbf{S}\in\mathbb{R}^3$). We employ a Partition-of-Unity (PU) spectral basis with $M$ non-negative basis functions using splines \cite{belcour_one--many_2023}:
    \begin{equation}
        \mathbf{B}\in\mathbb{R}^{K\times M},\qquad B_{\lambda m}\ge 0,\qquad \sum_{m=1}^{M} B_{\lambda m}=1\ \ \forall\,\lambda,
    \end{equation}
    and, if needed, an illuminant $\boldsymbol{\ell}\in\mathbb{R}^{K}$. We work with the illuminated basis
    \begin{equation}
        \tilde{\mathbf{B}}=\mathrm{diag}(\boldsymbol{\ell})\,\mathbf{B}\in\mathbb{R}^{K\times M}.
    \end{equation}
    For each basis function $m$, precompute its camera RGB and a luminance-like sum:
    \begin{equation}
        \mathbf{b}^{(m)}_{\!rgb}=\mathbf{Q}^\top \tilde{\mathbf{B}}_{(:,m)}\in\mathbb{R}^{3},\qquad
        S_m=\mathbf{1}^\top \mathbf{b}^{(m)}_{\!rgb}.
    \end{equation}
    Define ring coordinates $r_m=[\mathbf{b}^{(m)}_{\!rgb}]_1/S_m,\ g_m=[\mathbf{b}^{(m)}_{\!rgb}]_2/S_m$ and collect $\{(r_m,g_m)\}_{m=1}^M$. For a target pixel with $\mathbf{t}=(R,G,B)^\top$, set
    \begin{equation}
        S=R+G+B,\qquad r=\frac{R}{S},\qquad g=\frac{G}{S}.
    \end{equation}
    
    The chromaticities $\{(r_m,g_m)\}$ form a closed polygon in the chromaticity plane. Given a target chromaticity $(r,g)$, we express it as a convex combination of three ring vertices. For an oriented triple $(i,j,k)$, define
    \begin{equation}
        \mathbf{T}_{\triangle}=
    \begin{bmatrix}
        1 & 1 & 1\\
        r_i & r_j & r_k\\
        g_i & g_j & g_k
    \end{bmatrix},\qquad
    \boldsymbol{\xi}=
    \begin{bmatrix}
        1\\ r\\ g
    \end{bmatrix}.
    \end{equation}
    The triangle barycentric coordinates are $\boldsymbol{a}_T=\mathbf{T}_{\triangle}^{-1}\boldsymbol{\xi}$; a triangle contains the target iff $\boldsymbol{a}_T\ge 0$ and $\mathbf{1}^\top\boldsymbol{a}_T=1$. Valid triangles are pre-enumerated; one is selected (e.g., at random among valids) to anchor a local, well-conditioned representation of the target chromaticity. This identifies a minimal three-function basis that matches $(r,g)$ prior to luminance scaling.
    
    Fix a containing triangle $(i,j,k)$ and partition the ring into the triangle set $\mathbf{T}=\{i,j,k\}$ and the free set $\mathbf{F}=\{1,\dots,M\}\setminus\{i,j,k\}$. Collect the same affine embedding, $(1,r_m,g_m)$, on $\mathbf{T}$ and $\mathbf{F}$:
    \begin{equation}
    \mathbf{T}=
    \begin{bmatrix}
        1 & 1 & 1\\
        r_i & r_j & r_k\\
        g_i & g_j & g_k
    \end{bmatrix}\in\mathbb{R}^{3\times 3},\quad
    \mathbf{F}=
    \begin{bmatrix}
    \mathbf{1}^\top\\ \mathbf{r}^\top\\ \mathbf{g}^\top
    \end{bmatrix}\Big|_{\mathbf{F}}\in\mathbb{R}^{3\times (M-3)}.
    \end{equation}
    The geometry map
    \begin{equation}
    \mathbf{M}=\mathbf{T}^{-1}\mathbf{F}\in\mathbb{R}^{3\times (M-3)}
    \end{equation}
    encodes how increments on the free vertices must be counterbalanced on the triangle to preserve chromaticity exactly. Let $\boldsymbol{a}_F\in\mathbb{R}^{M-3}_{\ge 0}$ denote nonnegative degrees of freedom on $\mathbf{F}$. The generalized barycentric coordinates over the full ring are
    \begin{equation}
        \boldsymbol{a}=
    \begin{bmatrix}
        \boldsymbol{a}_T - \mathbf{M}\,\boldsymbol{a}_F\\[2pt]
        \boldsymbol{a}_F
    \end{bmatrix},\qquad
        \boldsymbol{a}\ge 0,\qquad \mathbf{1}^\top\boldsymbol{a}=1.
    \end{equation}
    By construction, $\boldsymbol{a}$ maps to $\boldsymbol{\xi}$ under the affine embedding, hence reproduces the target chromaticity $(r,g)$. Feasible upper bounds on each component of $\boldsymbol{a}_F$ are obtained from the nonnegativity of $\boldsymbol{a}_T-\mathbf{M}\boldsymbol{a}_F$, thereby parameterizing a convex family of spectra that share the same chromaticity. 
    
    Let $\mathbf{S}_{byL}=[S_1,\dots,S_M]^\top$. Enforce the target luminance $S$ by a diagonal rescaling:
    \begin{equation}
        \mathbf{w}=\alpha\,\boldsymbol{a}\oslash \mathbf{S}_{\!byL},\qquad
        \alpha\ \text{ s.t. }\ \mathbf{1}^\top\big(\mathbf{Q}^\top \tilde{\mathbf{B}}\mathbf{w}\big)=S.
    \end{equation}
    Because $\mathbf{1}^\top\boldsymbol{a}=1$, this reduces to $\alpha=S$, hence
    \begin{equation}
        \mathbf{w}=S\,(\boldsymbol{a}\oslash \mathbf{S}_{byL})
    \end{equation}
    The PU metamer is
    \begin{equation}
        \mathbf{S}_{\text{PU}}=\tilde{\mathbf{B}}\,\mathbf{w}\in\mathbb{R}^{K},\qquad
        \mathbf{Q}^\top\mathbf{S}_{\text{PU}}=\mathbf{t}
    \end{equation}
    
\begin{figure*}[ht]
    \centering
    \includegraphics[width=0.95\linewidth]{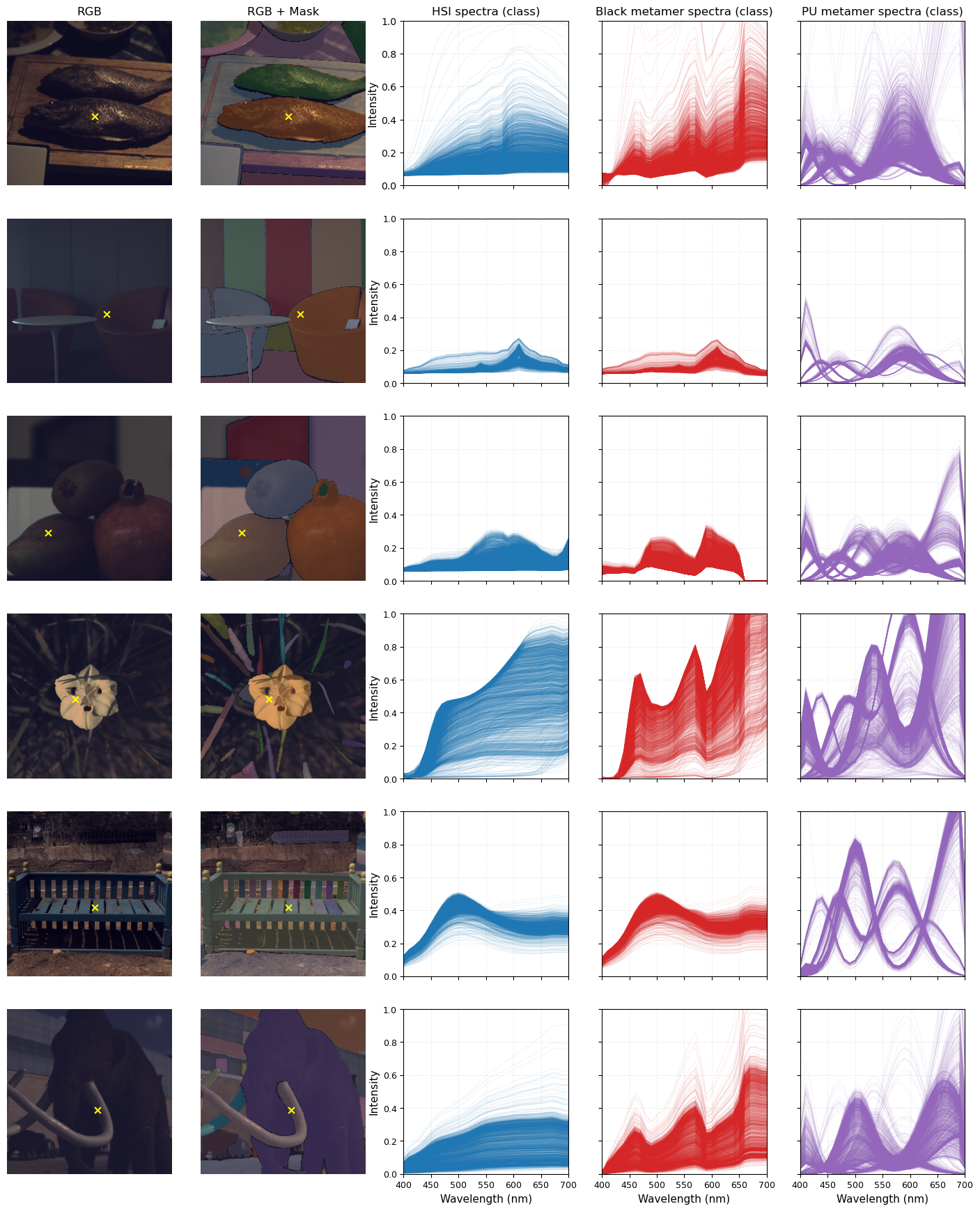}
    \caption{Metamer generation. From left to right: input RGB, segmentation mask, original HSI spectrum, black-metamer spectrum, and PU-basis metamer spectrum. All spectral profiles correspond to the same class indicated by the selected pixel on the mask.}
    \label{fig:metamers}
\end{figure*}

    Since $\mathbf{S}_{\text{PU}}=\tilde{\mathbf{B}}\mathbf{w}$ is a convex blend of smooth PU basis functions, spectra are inherently regularized. As we discretize the spectral integral as $\mathbf{t}=\mathbf{Q}^\top\mathbf{S}$, this can produce minor numerical overshoots beyond the physical $[0,1]$ range for $\mathbf{S}$. We therefore clip $\mathbf{S}$ to $[0,1]$. Empirically, this projection does not materially affect metamerism: RGB consistency remains within numerical tolerance, with reconstructions retaining very high fidelity (PSNR $> 70$ dB across tested samples). Figure \ref{fig:metamers} presents examples of texture-guided black metamers and PU-basis metamers across different scenes.
    
\section{Additional Results}
\begin{figure*}[ht]
    \centering
    \includegraphics[width=0.9\linewidth]{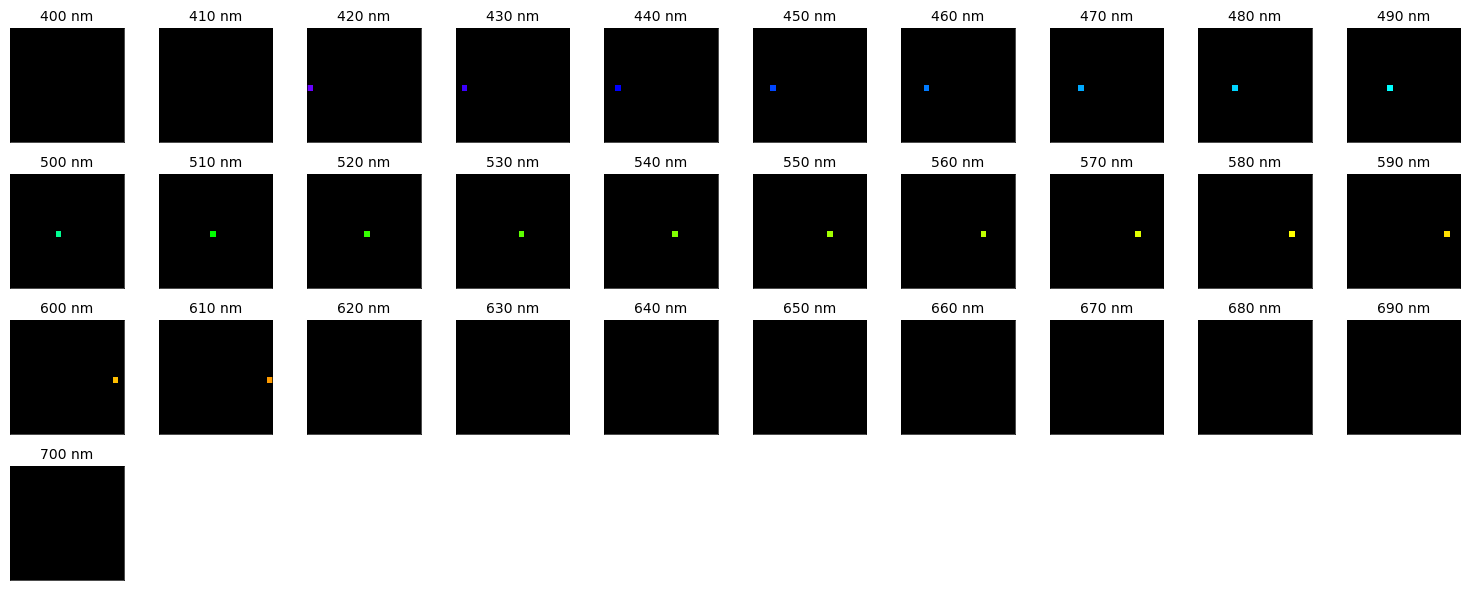}
    \caption{Grating PSF \cite{baek_compact_2017}. It shows the center 20 by 20 pixels of the actual 256 by 256 PSF to make it visible.}
    \label{fig:grating}
\end{figure*}

\begin{figure*}[ht]
    \centering
    \includegraphics[width=0.9\linewidth]{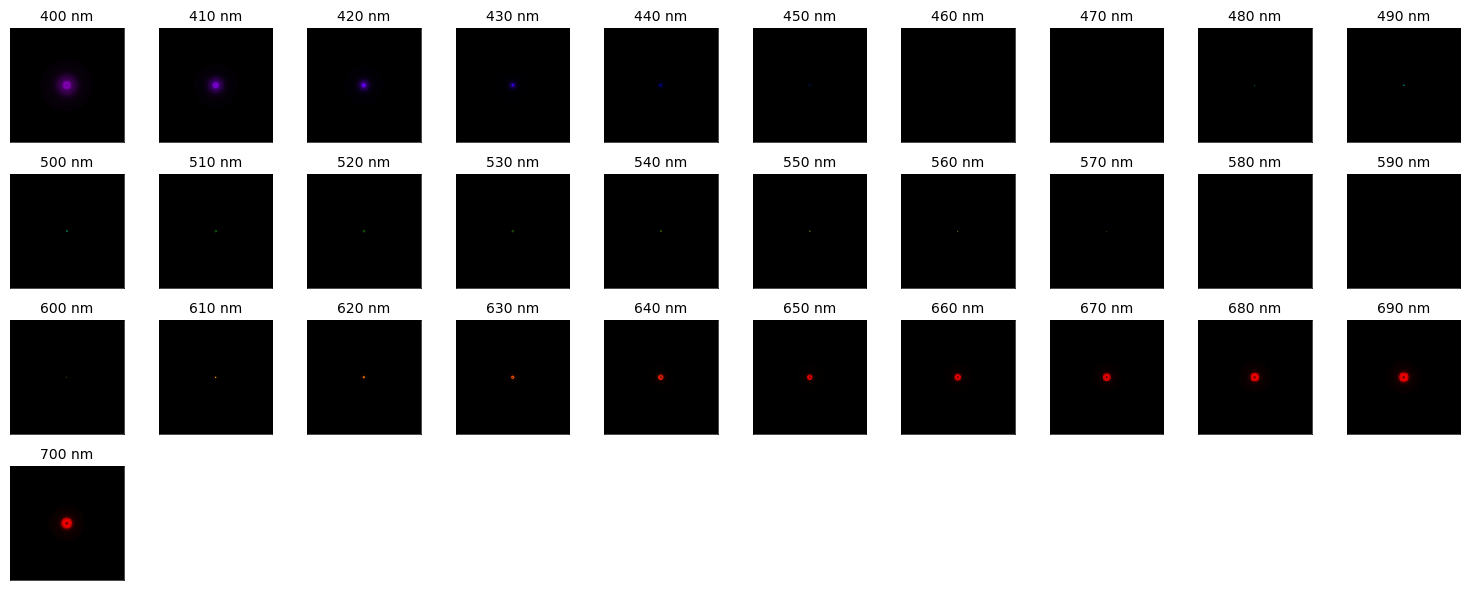}
    \caption{Gaussian PSF \cite{ichimura2021optical}.}
    \label{fig:gaussian}
\end{figure*}

\begin{figure*}[ht]
    \centering
    \includegraphics[width=0.9\linewidth]{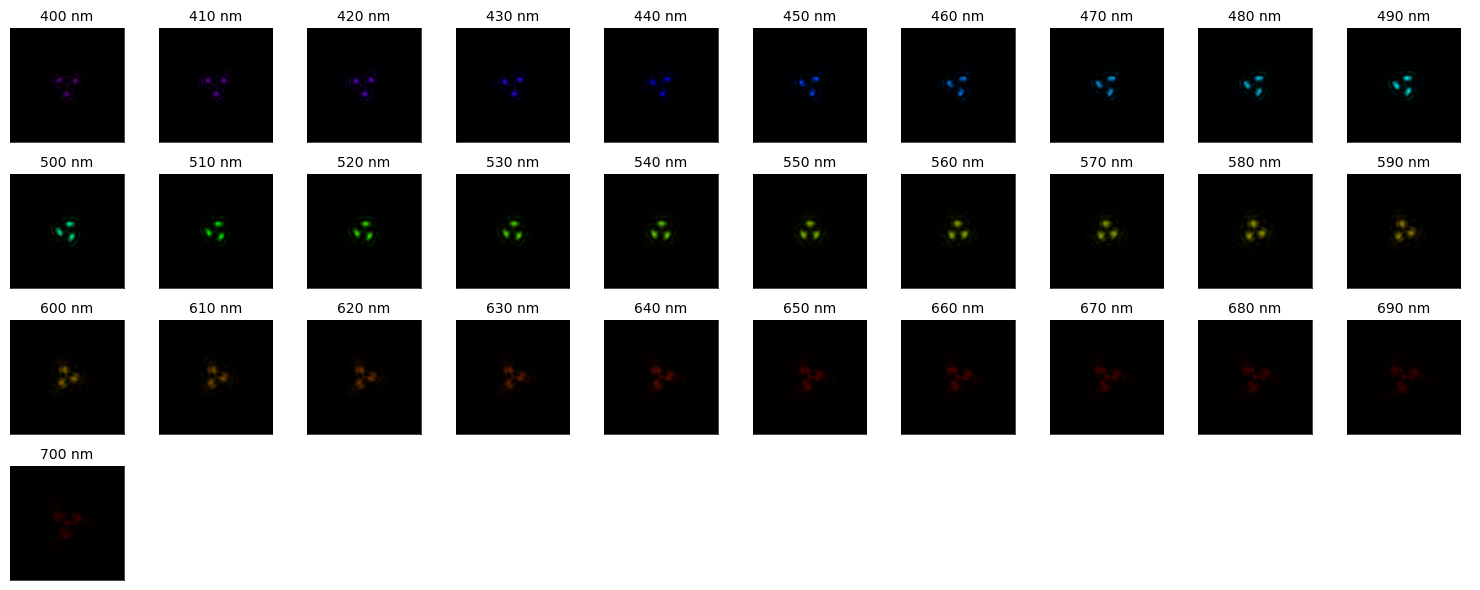}
    \caption{Rotational PSF \cite{jeon_compact_2019}.}
    \label{fig:rotational}
\end{figure*}

\begin{figure*}[ht]
    \centering
    \includegraphics[width=\linewidth]{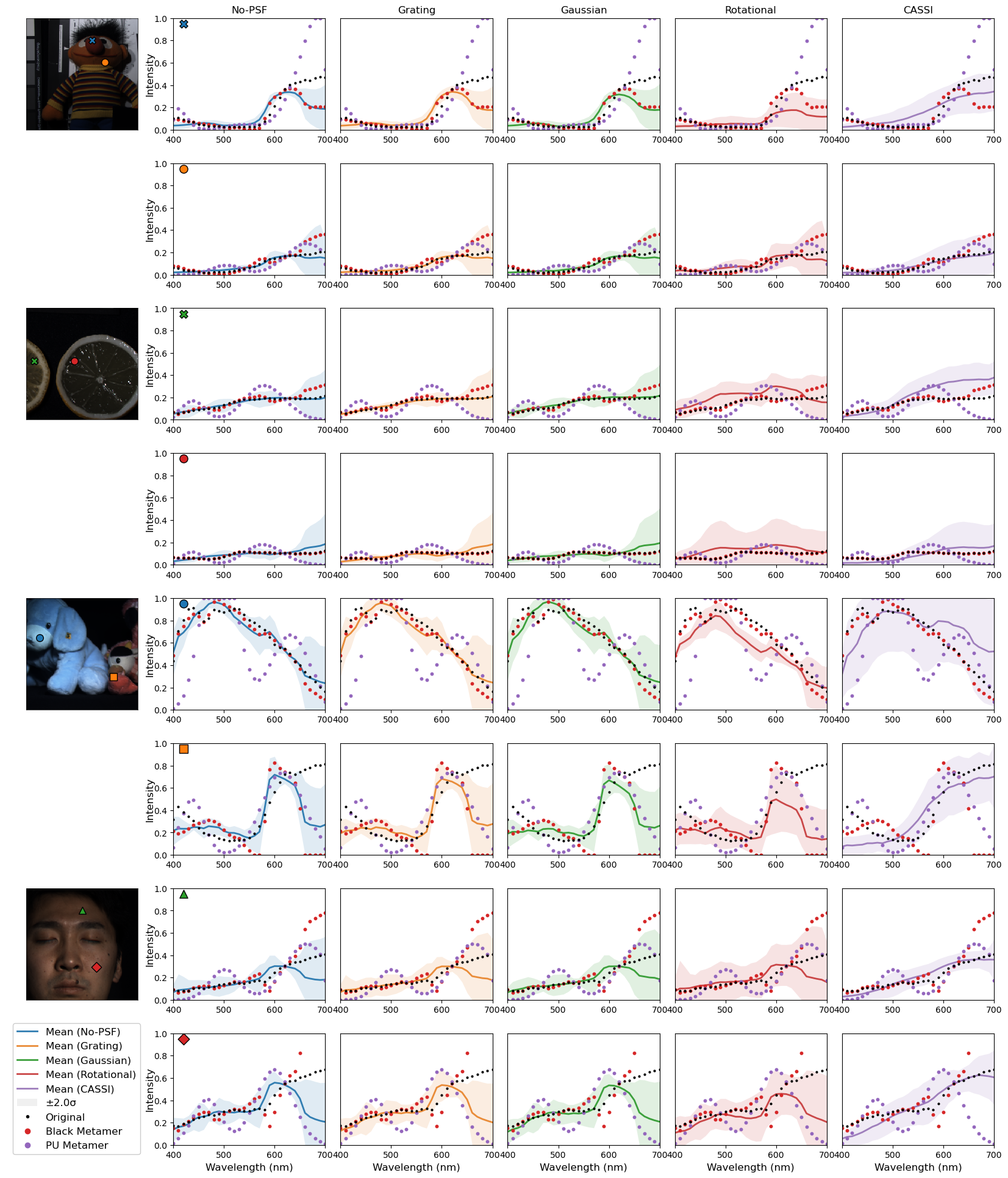}
    \caption{HSDiff results on CAVE dataset.}
    \label{fig:cave1}
\end{figure*}

\begin{figure*}[ht]
    \centering
    \includegraphics[width=\linewidth]{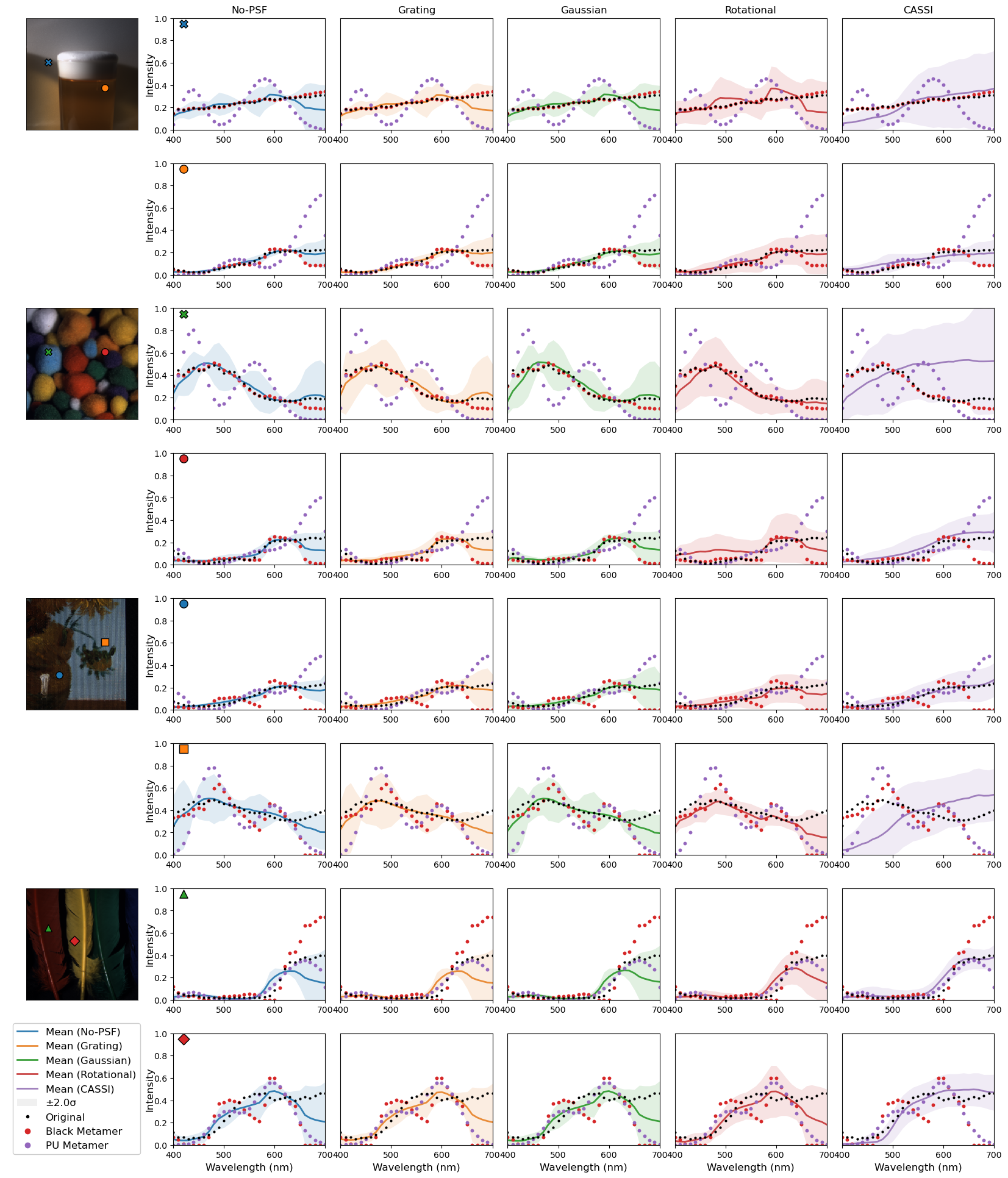}
    \caption{HSDiff results on CAVE dataset.}
    \label{fig:cave2}
\end{figure*}

\begin{figure*}[ht]
    \centering
    \includegraphics[width=\linewidth]{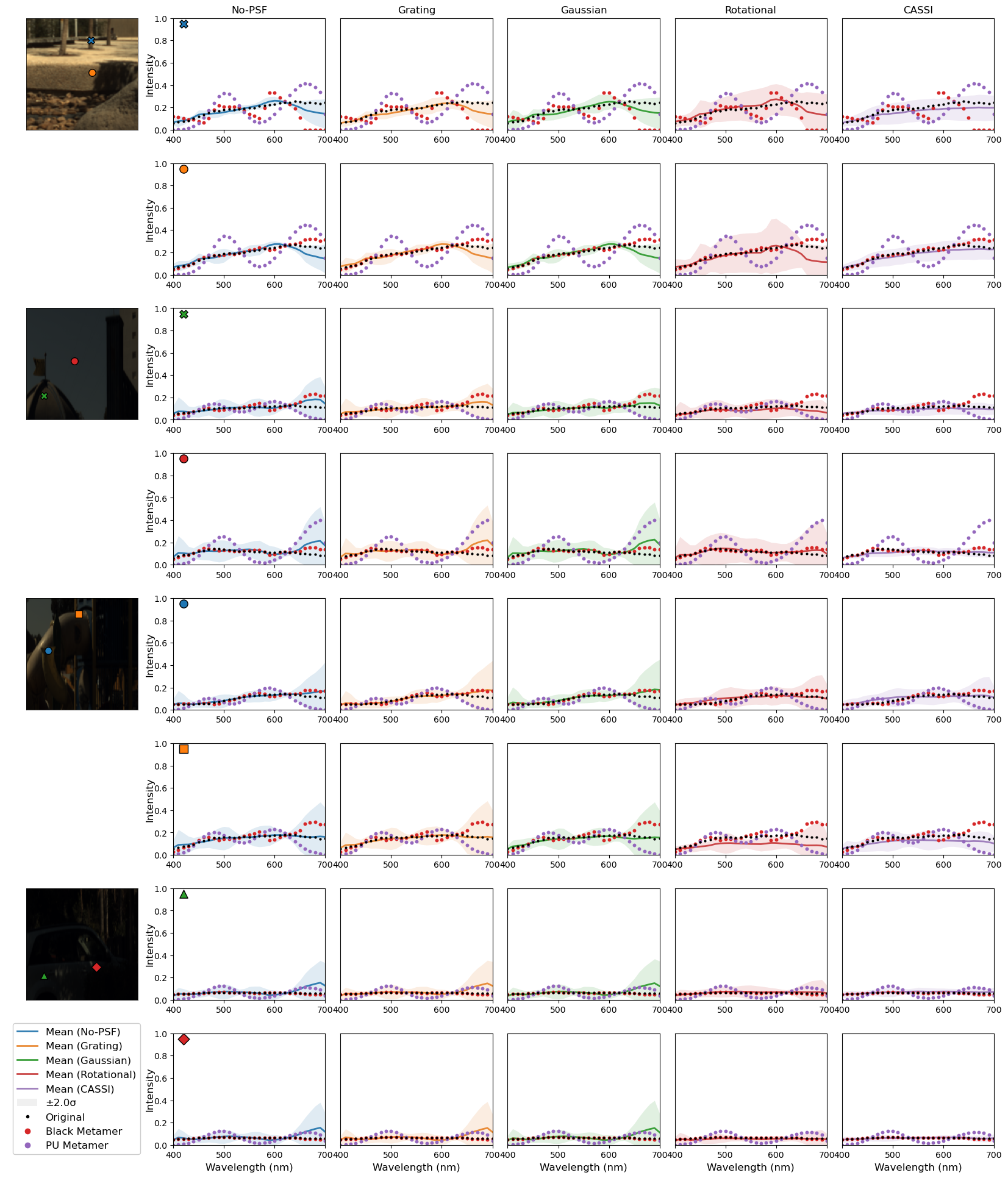}
    \caption{HSDiff results on ICVL dataset.}
    \label{fig:icvl1}
\end{figure*}

\begin{figure*}[ht]
    \centering
    \includegraphics[width=\linewidth]{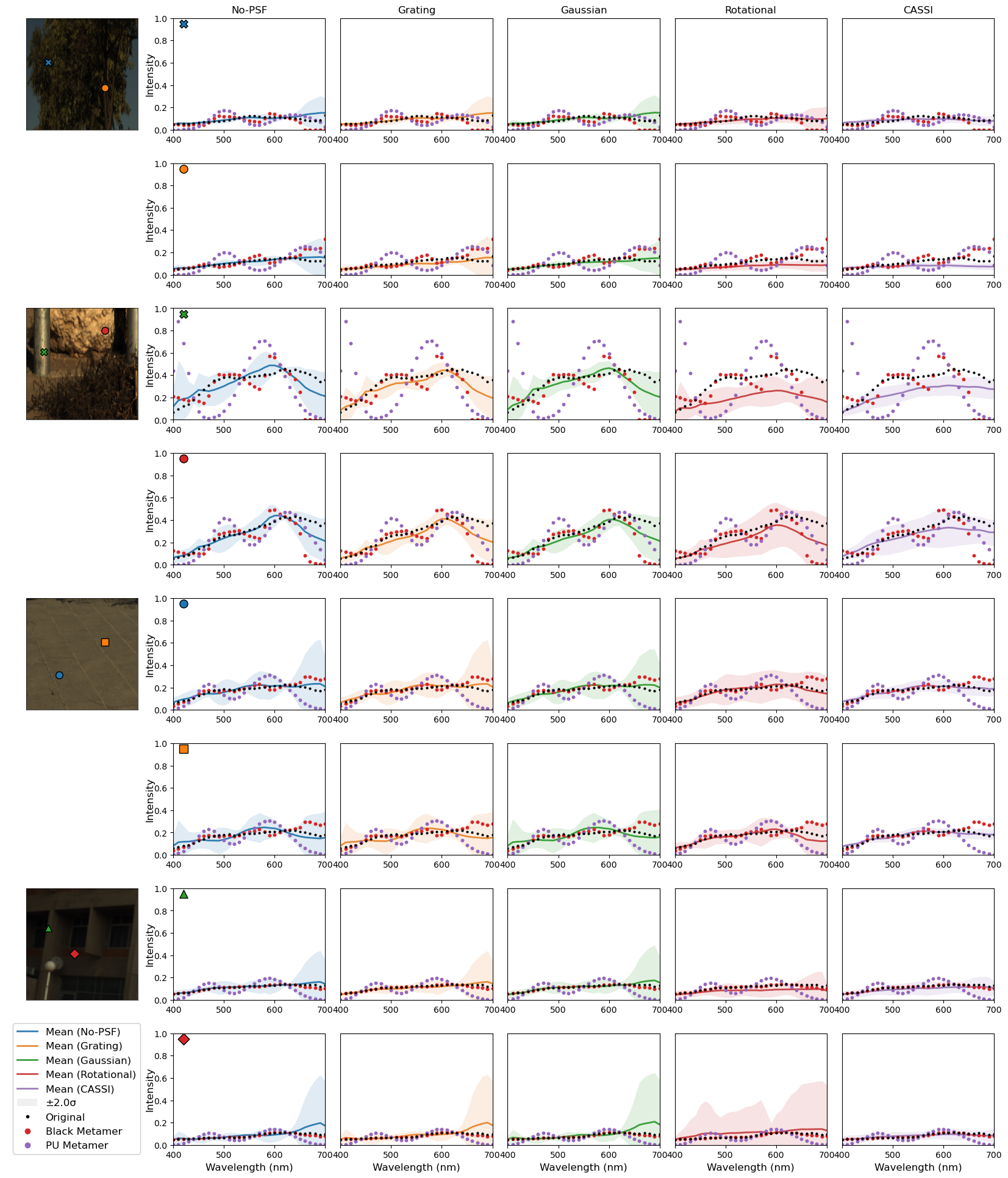}
    \caption{HSDiff results on ICVL dataset.}
    \label{fig:icvl2}
\end{figure*}

Figures \ref{fig:cave1}, \ref{fig:cave2}, \ref{fig:icvl1}, \ref{fig:icvl2} report additional HSDiff results across the five spectral encodings evaluated: None, Grating \cite{baek_compact_2017}(Figure \ref{fig:grating}), Gaussian \cite{ichimura2021optical} (Figure \ref{fig:gaussian}), Rotational \cite{jeon_compact_2019} (Figure \ref{fig:rotational}), and CASSI \cite{meng2020end}, in the CAVE \cite{yasuma_generalized_2010} and ICVL \cite{arad_sparse_2016} validation datasets. Overall, the None/Grating/Gaussian settings tend to be confidently wrong: their credible intervals are comparatively narrow yet miss ground truth more often, indicating poorer calibration. By contrast, Rotational and CASSI produce broader, better-calibrated intervals that capture a larger portion of metameric variability (though not perfectly). These trends suggest that optics-aware encodings, especially CASSI, yield uncertainty that more faithfully reflects task difficulty and, in turn, offer a clearer path toward improved system design.

\section{Ablation Studies}
\label{sec:sup_ablation}
We perform ablations on a full grid of the diffusion guided sampling hyperparameter $(\sigma_y,\nu,\lambda)$, for each of the forward modeling operators (No PSF, Grating PSF \cite{baek_compact_2017}, Gaussian PSF \cite{ichimura2021optical}, Rotational PSF \cite{jeon_compact_2019}, and CASSI \cite{meng2020end}) and for ICVL and CAVE validation datasets. Tables~\ref{tab:ablation_sigma_y}, \ref{tab:ablation_nu}, \ref{tab:ablation_lambda}, \ref{tab:ablation_sigma_y_cave}, \ref{tab:ablation_nu_cave}, \ref{tab:ablation_lambda_cave} report only 1D sweeps for No PSF operator on ICVL and CAVE as a compact summary. The complete ablation grids, including all parameter combinations and metrics, are provided as a CSV file attached to this submission.

\begin{table}[ht]
    \centering
    \caption{Ablation over likelihood noise scale $\sigma_y$ on ICVL (no PSF, EMA weights), 
    fixing $\nu = 1.0$ and $\lambda = 0.1$. Bold indicates best PSNR and best PICP 
    within the sweep.}
    \label{tab:ablation_sigma_y}
    \footnotesize
    \begin{tabular}{ccccc}
        \toprule
        $\sigma_y$ & PSNR$\uparrow$ & SAM$\downarrow$ & PICP$\uparrow$ \\
        \midrule
        0.001 & \textbf{30.524} & 10.006 & 0.9970 \\
        0.01  & 30.501          & 10.034 & 0.9971 \\
        0.1   & 30.375          & 10.082 & \textbf{0.9972} \\
        1.0   & 29.405          & 10.131 & 0.9957 \\
        2.0   & 28.704          & 10.161 & 0.9946 \\
        \bottomrule
    \end{tabular}
\end{table}

\begin{table}[ht]
    \centering
    \caption{Ablation over guidance weight $\nu$ on ICVL (no PSF, EMA weights),
    fixing $\sigma_y = 0.001$ and $\lambda = 0.1$. Bold indicates best PSNR 
    and best PICP within the sweep.}
    \label{tab:ablation_nu}
    \footnotesize
    \begin{tabular}{ccccc}
        \toprule
        $\nu$ & PSNR$\uparrow$ & SAM$\downarrow$ & PICP$\uparrow$ \\
        \midrule
        0.001 & 15.418 & 46.272 & 0.7460 \\
        0.01  & 12.898 & 59.480 & 0.3752 \\
        0.1   & 22.008 & 27.037 & 0.9644 \\
        1.0   & \textbf{30.524} & 10.006 & \textbf{0.9970} \\
        \bottomrule
    \end{tabular}
\end{table}

\begin{table}[ht]
    \centering
    \caption{Ablation over log-det weight $\lambda$ on ICVL (no PSF, EMA weights), fixing $\sigma_y = 0.001$ and $\nu = 1.0$. Bold indicates best PSNR 
    and best PICP within the sweep.}
    \label{tab:ablation_lambda}
    \footnotesize
    \begin{tabular}{ccccc}
        \toprule
        $\lambda$ & PSNR$\uparrow$ & SAM$\downarrow$ & PICP$\uparrow$ \\
        \midrule
        0.01 & 22.357 & 10.927 & 0.9961 \\
        0.1  & \textbf{30.524} & 10.006 & \textbf{0.9970} \\
        1.0  & 21.999 & 27.093 & 0.9643 \\
        \bottomrule
    \end{tabular}
\end{table}

\begin{table}[ht]
    \centering
    \caption{Ablation over likelihood noise scale $\sigma_y$ on CAVE (no PSF, EMA weights),
    fixing $\nu = 1.0$ and $\lambda = 0.1$. Bold indicates best PSNR and best PICP 
    within the sweep.}
    \label{tab:ablation_sigma_y_cave}
    \footnotesize
    \begin{tabular}{cccc}
        \toprule
        $\sigma_y$ & PSNR$\uparrow$ & SAM$\downarrow$ & PICP$\uparrow$ \\
        \midrule
        0.001 & \textbf{22.070} & 25.405 & 0.8771 \\
        0.01  & 22.022          & 25.591 & 0.8779 \\
        0.1   & 21.857          & 26.179 & 0.8781 \\
        1.0   & 21.131          & 27.946 & 0.8775 \\
        2.0   & 20.629          & 28.780 & \textbf{0.8855} \\
        \bottomrule
    \end{tabular}
\end{table}

\begin{table}[ht]
    \centering
    \caption{Ablation over guidance weight $\nu$ on CAVE (no PSF, EMA weights),
    fixing $\sigma_y = 0.001$ and $\lambda = 0.1$. Bold indicates best PSNR 
    and best PICP within the sweep.}
    \label{tab:ablation_nu_cave}
    \footnotesize
    \begin{tabular}{cccc}
        \toprule
        $\nu$ & PSNR$\uparrow$ & SAM$\downarrow$ & PICP$\uparrow$ \\
        \midrule
        0.001 & 14.649 & 49.277 & 0.5130 \\
        0.01  & 13.456 & 51.416 & 0.3613 \\
        0.1   & 17.543 & 36.718 & 0.7475 \\
        1.0   & \textbf{22.070} & 25.405 & \textbf{0.8771} \\
        2.0   & 21.696 & 28.345 & 0.8674 \\
        \bottomrule
    \end{tabular}
\end{table}

\begin{table}[ht]
    \centering
    \caption{Ablation over log-det weight $\lambda$ on CAVE (no PSF, EMA weights),
    fixing $\sigma_y = 0.001$ and $\nu = 1.0$. Bold indicates best PSNR 
    and best PICP within the sweep.}
    \label{tab:ablation_lambda_cave}
    \footnotesize
    \begin{tabular}{cccc}
        \toprule
        $\lambda$ & PSNR$\uparrow$ & SAM$\downarrow$ & PICP$\uparrow$ \\
        \midrule
        0.01 & 17.491 & 32.907 & \textbf{0.9071} \\
        0.1  & \textbf{22.070} & 25.405 & 0.8771 \\
        1.0  & 17.540 & 36.733 & 0.7485 \\
        \bottomrule
    \end{tabular}
\end{table}


\clearpage
\begin{algorithm*}[ht]
\caption{Guided EDM stochastic sampler with $\sigma(t)=t$ and $\mu(t)=1$}
\label{alg:guided_edm}
\begin{algorithmic}[1]
\Procedure{StochasticSampler}{$D_{\theta}(\mathbf{x};\sigma),\, \{t_i\}_{i=0}^{N},\, \{\gamma_i\}_{i=0}^{N-1},\, S_{\text{noise}}$}
\State sample $\mathbf{x}_{0}\sim\mathcal{N}(0,\,t_0^{2}\mathbf I)$
\For{$i\in\{0,\ldots,N-1\}$}
  \State sample $\varepsilon_i\sim\mathcal{N}(0,\,S_{\text{noise}}^{2}\mathbf I)$
  \State $\hat t_i \gets t_i + \gamma_i t_i$
  \Comment{$\displaystyle
  \gamma_i=
  \begin{cases}
  \min\!\left(\frac{S_{\text{churn}}}{N},\,\sqrt{2}-1\right), & t_i\in [S_{\min},S_{\max}]\\[2pt]
  0, & \text{otherwise}
  \end{cases}$}
  \State $\hat{\mathbf{x}}_i \gets \mathbf{x}_i + \sqrt{\hat t_i^{\,2}-t_i^{\,2}}\;\boldsymbol{\varepsilon}_i$
  \Comment{Add noise to move from $t_i$ to $\hat t_i$}
  \State $w_i \gets \lambda / (\sigma_y + \hat t_i^2 \ \nu)$   
  \State ${\color{blue}\mathbf{s}_l \gets w_i \,\nabla_{\hat{\mathbf{x}}_i} \bigl\|\,\mathbf{y} - \mathcal{A}(D_{\theta}(\hat{\mathbf{x}}_i;\hat t_i))\bigr\|_2^{2}}$
  \State $\mathbf{d}_i \gets \big(\hat{\mathbf{x}}_i - D_{\theta}(\hat{\mathbf{x}}_i;\hat t_i)\big) /\hat t_i + {\color{blue} \hat t_i s_l}$
  \Comment{Evaluate $\mathrm d \mathbf{x}/\mathrm d t$ at $\hat t_i$}
  \State $\mathbf{x}_{i+1} \gets \hat{\mathbf{x}}_i + (t_{i+1}-t_i)\, \mathbf{d}_i$
  \Comment{Euler step from $\hat t_i$ to $t_{i+1}$}
  \If{$t_{i+1}\neq 0$}
  \State $w_{i+1} \gets \lambda / (\sigma_y + \hat t_{i+1}^2 \ \nu)$ 
   \State ${\color{blue}\mathbf{s}'_l \gets w_{i+1} \,\nabla_{\hat{\mathbf{x}}_{i+1}} \bigl\|\,\mathbf{y} - \mathcal{A}(D_{\theta}(\hat{\mathbf{x}}_{i+1};\hat t_{i+1}))\bigr\|_2^{2}}$
    \State $\mathbf{d}'_i \gets \big(\mathbf{x}_{i+1}-D_{\theta}(\mathbf{x}_{i+1};t_{i+1})\big) /\hat t_{i+1} + {\color{blue} \hat t_{i+1} s'_l}$
    \State $\mathbf{x}_{i+1} \gets \hat{\mathbf{x}}_i + (t_{i+1}-t_i)\!\left(\tfrac{1}{2}\mathbf{d}_i+\tfrac{1}{2}\mathbf{d}'_i\right)$
    \Comment{Apply 2\textsuperscript{nd}-order (Heun) correction}
  \EndIf
\EndFor
\State \Return $x_N$
\EndProcedure
\end{algorithmic}
\end{algorithm*}

\clearpage

\clearpage

\clearpage


\end{document}